%% file: main.tex
\documentclass{article}

\usepackage[preprint]{neurips_2024}

\usepackage{subcaption}
\usepackage{diagbox}
\usepackage{supertabular}
\usepackage{colortbl}
\usepackage{bbding}
\usepackage[utf8]{inputenc} 
\usepackage[T1]{fontenc}    
\usepackage{microtype}      
\usepackage{xcolor}         
\usepackage{xspace}
\usepackage{amsmath,amssymb,amsfonts,dsfont,pifont,bm,bbm,mathrsfs,mathtools,nicefrac}
\usepackage{algorithm,algpseudocode,listings}
\usepackage{wrapfig}
\usepackage{booktabs,multirow,adjustbox,diagbox,threeparttable,makecell}
\definecolor{citeblue}{HTML}{0071bc}
\definecolor{textpurple}{RGB}{135,89,201}
\definecolor{Gray}{gray}{0.90}
\usepackage[pagebackref=true,breaklinks=true,colorlinks=true,citecolor=citeblue,bookmarks=false]{hyperref}
\usepackage{url}            
\usepackage{cleveref}  
\usepackage{lipsum}
\usepackage{graphicx}
\usepackage{natbib}
\setcitestyle{numbers,square}

\crefname{section}{Sec.}{Secs.}
\Crefname{section}{Section}{Sections}
\crefname{appendix}{Appendix}{Appendixes}
\crefname{table}{Tab.}{Tabs.}
\Crefname{table}{Table}{Tables}
\crefname{figure}{Fig.}{Figs.}
\Crefname{figure}{Figure}{Figures}
\crefname{equation}{Eq.}{Eqs.}
\Crefname{equation}{Equation}{Equations}
\newcommand{\method}{M2-omni\xspace}

\title{\method: Advancing Omni-MLLM for Comprehensive Modality Support with Competitive Performance}

\author{%
  Qingpei~Guo\thanks{Corresponding author.}\quad
  Kaiyou~Song\thanks{Equal Contributors.}\quad
  Zipeng~Feng\footnotemark[2]\quad
  Ziping~Ma{\footnotemark[2]} \quad
  Qinglong~Zhang \quad
  Sirui~Gao \\
  \textbf{Xuzheng~Yu} \quad
  \textbf{Yunxiao~Sun}\quad
  \textbf{Tai-Wei~Chang}\quad
  \textbf{Jingdong~Chen}\quad
  \textbf{Ming~Yang} \quad
  \textbf{Jun~Zhou}\\[1pt]
  Ant Group\\
  \texttt{\{qingpei.gqp,jingdongchen.cjd\}@antgroup.com}\\
  \enskip \\
}

\begin{document}

\maketitle
\input{sections/0.abstract.tex}

\input{sections/1.introduction.tex}

\input{sections/2.overall_Architecture.tex}

\input{sections/3.approach.tex}
\input{sections/4.experiment.tex}
\input{sections/5.conclusion.tex}

\input{sections/6.ref.tex}

\newpage
\input{sections/7.appendix.tex}

\end{document}

%% file: sections/0.abstract.tex
\begin{figure}[h]
    \centering
    \includegraphics[width=0.9\linewidth]{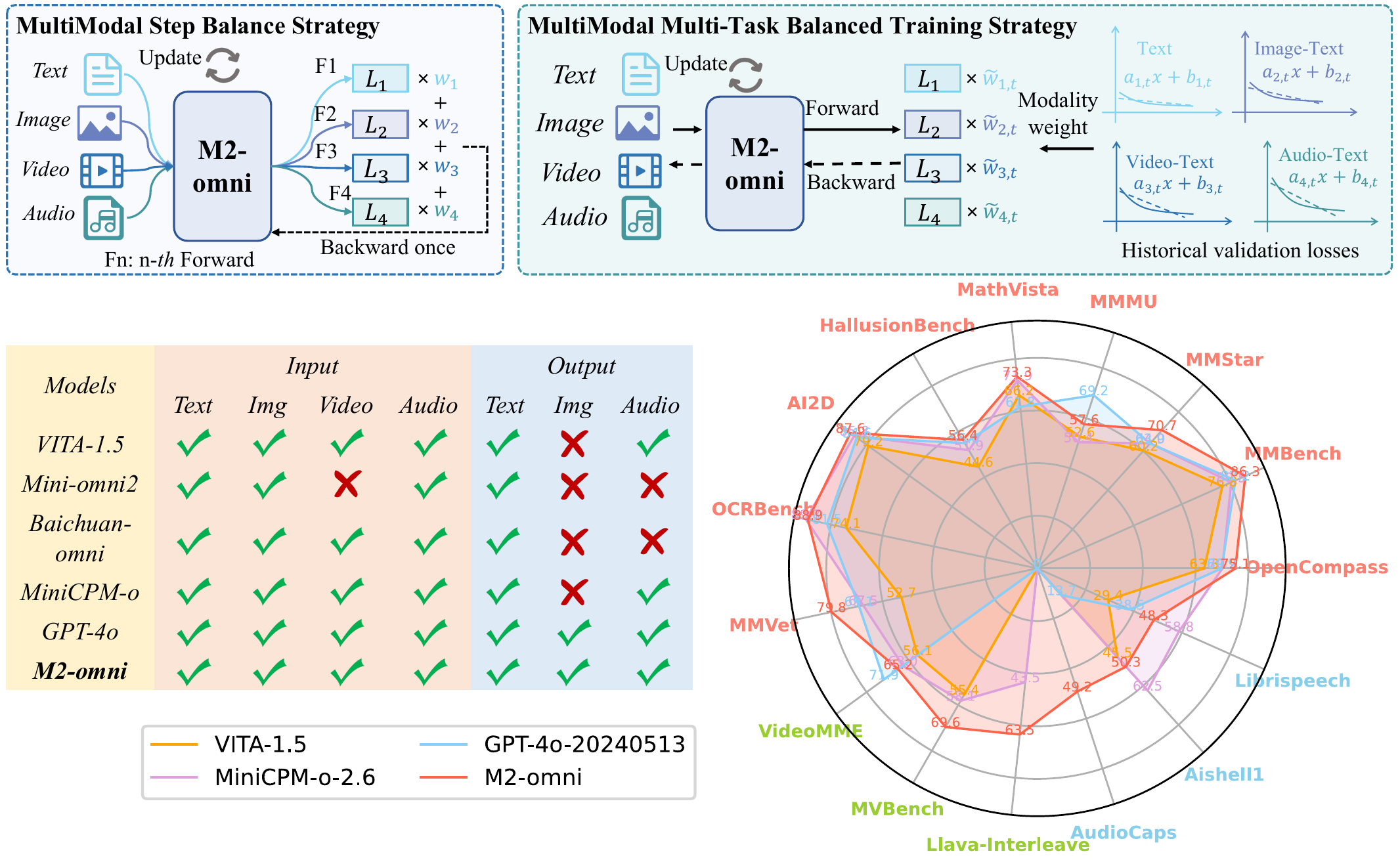}
    \caption{
    \textbf{Overall illustration of \method.}
    (Top) \method employs a multi-stage training with progressively modality alignment and multimodal multi-task balanced training strategy to achieve the optimal performance of each modality.
    (Left-bottom) \method supports as many modalities and tasks as other omni-MLLMs combined.
    (Right-bottom) \method achieves competitive performances on a broad range of multimodal tasks among its omni-MLLM counterparts. Note that the values on Librispeech~\cite{Librispeech} and Aishll1~\cite{AISHELL1} are taken the reciprocal for better visualization, and the results on Librispeech~\cite{Librispeech} and Aishll1~\cite{AISHELL1} of GPT-4o (GPT-4o-Realtime) are taken from \cite{yao2024minicpm}.
    More comprehensive results can be found in \cref{sec:exp}.
    }
    \label{fig-modality_support}
\end{figure}

\begin{abstract}

We present \method, a cutting-edge, open-source omni-MLLM that achieves competitive performance to GPT-4o.  \method employs a unified multimodal sequence modeling framework, which empowers Large Language Models (LLMs) to acquire comprehensive cross-modal understanding and generation capabilities.  Specifically, \method can process arbitrary combinations of audio, video, image, and text modalities as input, generating multimodal sequences interleaving with audio, image, or text outputs, thereby enabling an advanced and interactive real-time experience.  The training of such an omni-MLLM is challenged by significant disparities in data quantity and convergence rates across modalities. To address these challenges, we propose a step balance strategy during pre-training to handle the quantity disparities in modality-specific data.  Additionally,  a dynamically adaptive balance strategy is introduced during the instruction tuning stage to synchronize the modality-wise training progress, ensuring optimal convergence.  Notably, we prioritize preserving strong performance on pure text tasks to maintain the robustness of \method's language understanding capability throughout the training process. To our best knowledge, \method is currently a very competitive open-source model to GPT-4o, characterized by its comprehensive modality and task support, as well as its exceptional performance. We expect \method will advance the development of omni-MLLMs, thus facilitating future research in this domain.

\end{abstract}

%% file: sections/1.introduction.tex
\section{Introduction}\label{sec:intro}

The recent breakthroughs in Large Language Models (LLMs)~\cite{openai2023gpt4, llama3_2024, reid2024gemini1_5} have significantly accelerated the development of Multimodal LLMs (MLLMs)~\cite{openai2024gpt4ocard, llama32, internvl_2024, llava-next_2024, qwen2-vl_2024}.
Omni-MLLM expands MLLM's capabilities by incorporating additional non-linguistic modalities such as video, audio, and others, thereby facilitating a more comprehensive and multidimensional understanding of the world. A prominent example of Omni-MLLM is GPT-4o~\cite{openai2024gpt4ocard}, which has demonstrated remarkable multimodal processing capabilities.  GPT-4o features a novel, unified framework that can process arbitrary combinations of text, audio, image, and video inputs and generate outputs across multiple modalities, including text, audio, and image. Consequently, this enables more natural and intuitive human-computer interaction, marking a crucial milestone on the path toward AGI. The research community has been actively enriching Omni-MLLM by incorporating additional modalities and task support in recent years~\cite{baichuan-omni,emu3,deepseek_janus,muse_vl, mini_omni,mini_omni2, vita,fu2025vita,yao2024minicpm}. However, existing works fall short of matching GPT-4o's comprehensive modality support and task versatility.  Current omni-MLLM models are constrained by their limited support for either audio modalities~\cite{baichuan-omni, emu3, deepseek_janus,muse_vl}, which impede real-time interaction, or visual generation tasks~\cite{mini_omni, vita}, thereby restricting their applicability in visual applications.  To advance current Omni-MLLM towards a more sophisticated GPT-4o level counterpart, three primary challenges must be addressed: (1) developing a unified framework for multimodal understanding and generation tasks, (2) designing training strategies and pipelines that prevent performance degradation across all modalities, and (3) a detailed training protocol to achieve exceptional performance.

A fundamental challenge in building a unified multimodal framework arises from the disparate representational spaces required for understanding and generation tasks. These differences, as noted by~\cite{deepseek_janus}, often lead to performance degradation in a shared model, particularly in terms of task-specific accuracy and generalization. In response to this challenge, we propose a unified modeling framework designed to effectively integrate multimodal understanding and generation tasks. Our framework features modality-specific processing pathways and utilizes a multi-stage training approach with progressive modality alignment, thereby mitigating interference between modalities and tasks. Specifically, for image generation tasks, building upon the works~\cite{li2023textbind, unifiedmllm, wang2024modaverse}, we employ textual descriptions as an intermediate representation, thereby circumventing the need for direct alignment of latent image features. For speech generation, leverage the model to predict discrete audio tokens enabling real-time, streaming audio synthesis with minimal impact on the performance of other modality branches.

A particular challenge in training Omni-MLLMs is maintaining consistent performance across all modalities when involving many modalities or tasks. This performance degradation often arises from significant disparities in data quantity and convergence rates across different tasks. In this work, we introduce a step balance strategy during pre-training. At each training iteration, a mini-batch comprising samples from each modality is sampled to maintain a balanced representation across all modalities, thereby mitigating bias caused by imbalanced data distribution among modalities. Furthermore, during the instruction tuning stage, we employ a dynamically adaptive balance strategy to regulate the convergence rates across modalities. The underlying rationale is that if one modality exhibits a slower convergence rate, it should be assigned a smaller gradient weight in model updates, thereby allowing faster-converging modalities to take precedence. Conversely, if a modality exhibits a faster convergence rate, it should be assigned a larger gradient weight. By adopting this balanced strategy, we can achieve enhanced performance across all modalities within the framework of omni-modal learning.

Currently,  open-source omni-MLLMs still exhibit a significant performance discrepancy in multimodal understanding compared to GPT-4o, which limits their wide application in industrial scenarios. Compared to other omni-MLLM efforts, our \method achieves state-of-the-art (SOTA) performance among publicly available omni-MLLMs. Specifically, our largest model, \method-72B, achieved an average score of 75.1 on the OpenCompass benchmark for vision-and-text tasks. This score even surpasses the performance of many vision-language specific MLLMs and proprietary commercial models.  Furthermore, we are publicly releasing the comprehensive training details, including data configurations and training procedures to develop \method. This detailed resource is intended to serve as a valuable guide for the community,  fostering research and development aimed at bridging the performance gap between open-source omni-MLLMs and GPT-4o.

To summarize, our contributions are as follows:

\begin{itemize}
    \item We propose \method, an advanced MLLM that demonstrates competitive performance among publicly available omni-MLLMs. \method represents a milestone in comprehensive modality and task support, narrowing the performance gap with proprietary models like GPT-4o. We publicly release the \method, as well as its comprehensive training details, including data configurations and training procedures.
    \item We propose a unified multimodal modeling framework that leverages a multi-stage training approach to achieve progressive modality alignment, enabling the effective integration of multimodal understanding and generation tasks.
    By implementing modality-specific processing pathways and innovative techniques such as text-based image generation and discrete token prediction-based speech generation, we minimize cross-modal interference while achieving comprehensive audio, video, image, and textual understanding and generation capabilities.
    \item To alleviate performance degradation when integrating multiple modalities, we propose a step balance strategy for pre-training and a dynamically adaptive balance strategy for SFT. This approach mitigates the impact due to significant variations in data volume and convergence rates across heterogeneous multimodal tasks.
\end{itemize}

%% file: sections/2.overall_Architecture.tex
\section{Unified Framework for MultiModal Understanding and Generation}\label{sec:overall_architecture}

\begin{figure}[t]
    \centering
    \includegraphics[width=1.0\linewidth]{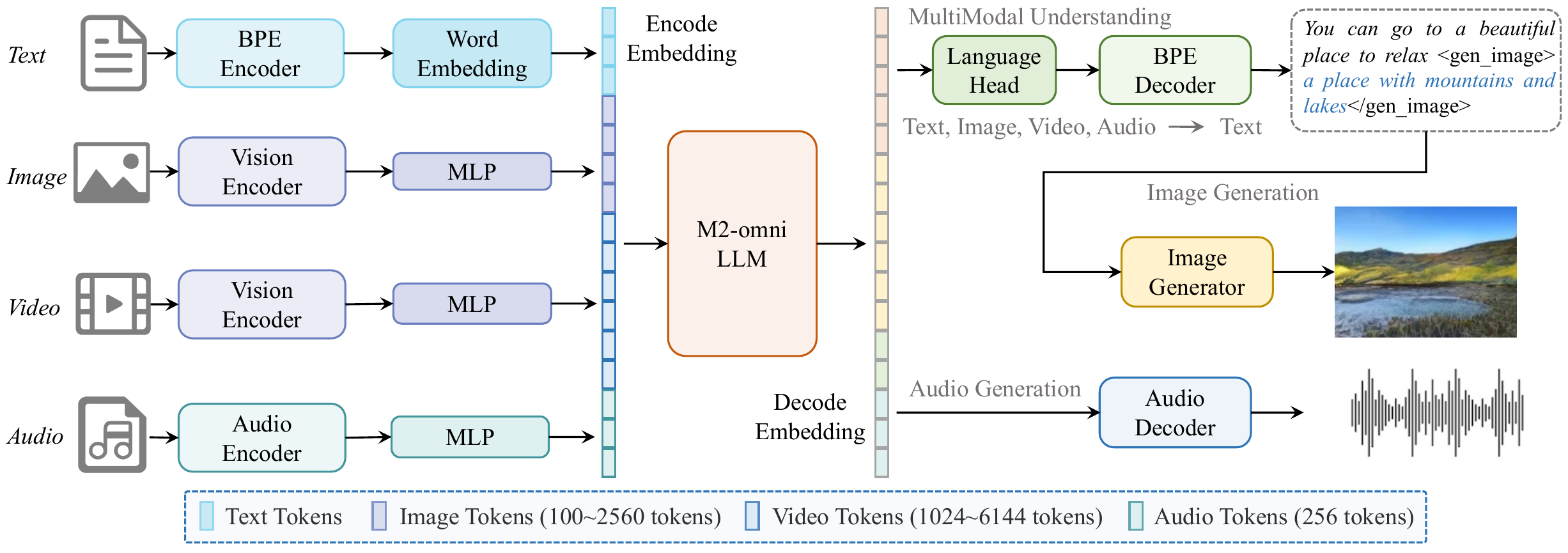}
    \caption{
    \textbf{Overall architecture of \method.} \method can process arbitrary combinations of text, image, video, and audio modalities as input, generating multimodal sequences interleaving with text, image, or audio outputs.
    }
    \label{fig-architecture}
\end{figure}


\subsection{Overall Architecture}\label{subsec:app_architecture}


We aim to build a unified framework that simultaneously supports multimodal understanding and generation tasks, while minimizing interference between different modality tasks through decoupled architecture design. Our encoding procedure is inspired by the design of UNIFIED-IO2~\cite{unified_io2}, which utilizes a modality-aware encoder to map diverse inputs (such as images, text, audio, and video) into a shared token representation space. Previous studies, such as Janus~\cite{Janus}, have demonstrated that multimodal understanding and generation tasks can interfere with each other, mainly due to the disparate levels of information granularity required for image understanding and generation.  In contrast to Janus~\cite{Janus}, which employs separate pathways for visual encoding, we leverage textual descriptions as an intermediate representation for image generation tasks, effectively bypassing the need for direct alignment of latent image features. For speech generation, we adopt a discrete token prediction-based approach, which enables real-time, streaming audio synthesis while minimizing the impact on the performance of other modality branches. Figure 2 illustrates the overall architecture of the proposed model. We will elaborate on the details of each module below.


\textbf{Vision Encoder.}
In \method, the vision encoder extracts representations from images or whole videos. We utilize a NaViT~\cite{navit_2024}  as the vision encoder, capable of processing videos and images of arbitrary resolution. To reduce the length of visual tokens, we concatenate adjacent $2\times2$ tokens into a single token and use an MLP to reduce the dimension to the original dimension, thereby downsampling the visual representation.

\textbf{Audio Encoder.}
We utilize the SAN-M~\cite{gao2020san, gao2022paraformer} encoder to extract audio tokens. Subsequently, we apply 1x3 average pooling to the audio encoder's output, aggregating every three adjacent tokens into a single token, which reduces the overall number of audio tokens. To accommodate the variability in audio token sequence lengths, we pad the compressed audio sequence with special \texttt{<audio\_pad>}  tokens, thereby ensuring that all sequences conform to a uniform length.


\textbf{M2-omni LLM.} The M2-omni LLM integrates the multimodal information and outputs the decoder embedding for unified multimodal understanding and generation. Our M2-omni LLM is initialized with pre-trained weights from the Llama3~\cite{llama_2023, llama3_2024} series, specifically Llama3.1-8B or Llama3.3-70B. To facilitate unified positional encoding across textual, image, video, and audio modalities, and to enable the model to generalize to longer sequences during inference, we substitute the original 1D-RoPE~\cite{su2024roformer} in Llama with M-RoPE~\cite{qwen2-vl_2024}.

\textbf{Image Generator.} To decouple the representation spaces of generation and understanding, building upon the insights from~\cite{li2023textbind, unifiedmllm, wang2024modaverse}, we utilize textual descriptions as an intermediate representation for image generation. During training, we warp the image captions with two special tokens, i.e. \texttt{<gen\_image>} and \texttt{</gen\_image>},  allowing the model to generate textual descriptions for image generation in a flexible and unconstrained manner. At inference time, the \method LLM generates the textual description, and the generated captions enclosed by the two special tokens are utilized as the textual condition for image generation. We employ an offline Stable Diffusion (SD) model~\cite{sd_2022} as the image generator.

\textbf{Audio Decoder.}
Inspired by the approaches in ~\cite{MinMo, mini_omni2}, we utilize the M2-omni LLM to predict discrete audio tokens for speech generation in an end-to-end style. The predicted discrete audio tokens are then fed into the pretrained CosyVoice~\cite{du2024cosyvoice} flow matching and vocoder model to generate audio streams. Given the similarity in form between audio discrete tokens and language tokens, we can repurpose the M2-omni LLM's model structure to facilitate audio generation tasks, thereby enabling compatibility with multimodal understanding tasks.


\cref{tab-model_config} illustrates the detailed model configuration of \method, and \cref{{fig-template}} demonstrates the data templates for image, video, and audio.

\begin{table}[t]
\centering\footnotesize
\caption{
\textbf{Detailed pre-trained model configuration of \method.}
}
\setlength{\tabcolsep}{3pt}
\begin{tabular}{c|c|c|c|c|c}
\toprule
Model Name & \#Param & Vision Encoder & Audio Encoder & LLM & SD \\
\midrule
\method-9B & 8.8B & \multirow{2}{*}{ViT-600M~\cite{qwen2-vl_2024}} & \multirow{2}{*}{paraformer-zh~\cite{gao2022paraformer}} & Llama3.1-8B~\cite{llama3_2024} & \multirow{2}{*}{SD-3-medium~\cite{esser2403scaling}} \\
\method-72B & 71.8B & & & Llama3.3-70B~\cite{llama3_2024} & \\
\bottomrule
\end{tabular}
\label{tab-model_config}
\end{table}

\begin{figure}[t]
    \centering
    \includegraphics[width=0.9\linewidth]{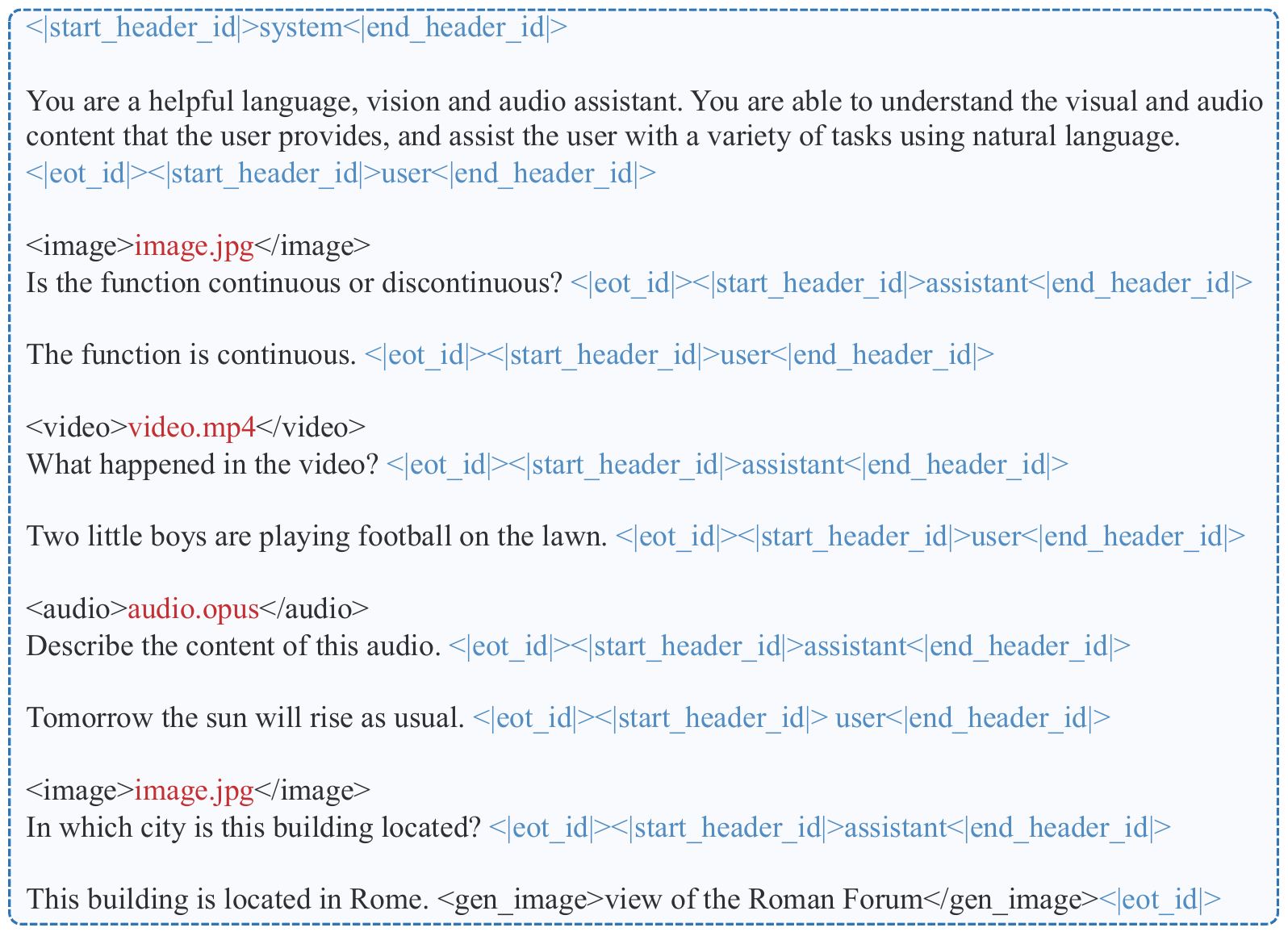}
    \caption{
    \textbf{Illustration of the templates of image, video, and audio.}
    }
    \label{fig-template}
\end{figure}

\subsection{Multi-Stage Training with Progressively Modality Alignment}\label{subsec:app_training}

\begin{figure}[t]
    \centering
    \includegraphics[width=1.0\linewidth]{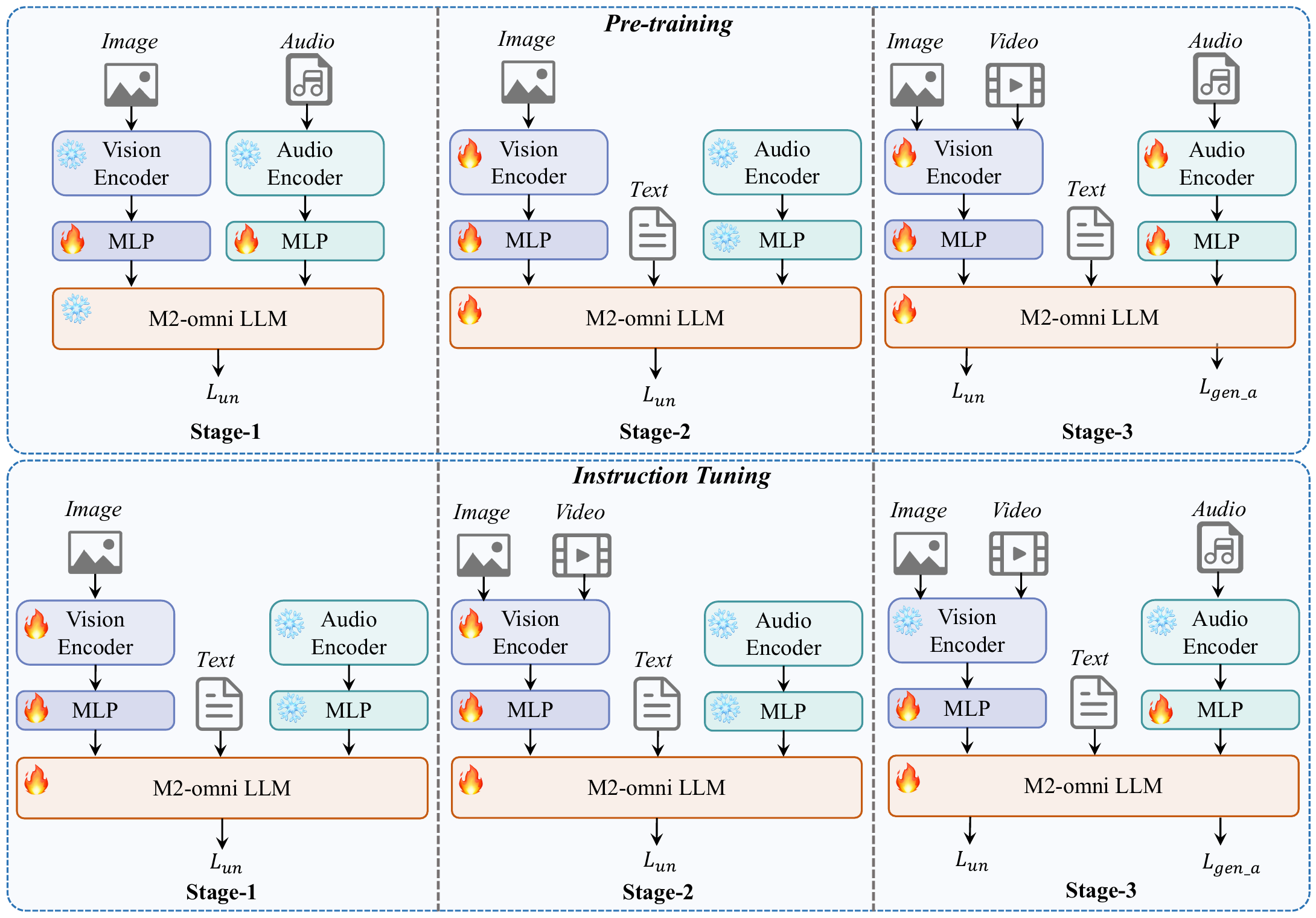}
    \caption{
    \textbf{Illustration of the training pipeline of \method.}
    Both the pre-training and the instruction tuning contain three stages, designed to progressively absorb knowledge from more modalities and ensure the model's optimal performance on all modalities and tasks. $L_{un}$ and $L_{gen\_a}$ denote understanding and audio generation loss, respectively.
    }
    \label{fig-pretrain_pipeline}
\end{figure}

Given a multimodal dataset, we employ modality-aware encoders to project diverse modality inputs, including images, text, audio, and video, into a unified token representation space. Formally, the input multimodal sequences are denoted as  $x = (x_1, \cdots, x_\ell)$, where $\ell$ represents the length of the sequence, and each $x_i$ corresponds to a modality input token (e.g., image, text, audio, or video).  In particular, we model the joint probability distribution of the multimodal sequence in an autoregressive manner, where each token is conditioned on the previous tokens, as shown in the following equation:
\begin{equation}
\log p_{\theta} (x) = \sum_{i = s}^{\ell-1} \log p_{\theta}(x_{i+1} | x_{0}, \dots, x_{i}),
\end{equation}
Notably, $s$ denotes the start index of discrete output tokens and only discrete output tokens $x_{>s}$ are considered as the modeling targets, $\theta$ denotes the parameters of the model.
We introduce a multi-stage training framework that progressively achieves modality alignment by incrementally incorporating knowledge from multiple modalities. As shown in \cref{fig-pretrain_pipeline}, the overall training procedure of our proposed \method consists of three primary stages: pre-training, instruction tuning, and alignment tuning.  Both the pre-training and instruction tuning stages are further divided into three sub-stages, each designed to incrementally incorporate additional modalities. The training hyperparameters and configurations are summarized in \cref{tab:app_train_hyperparameter}.

\subsubsubsection{\textbf{Pre-training}}

The pre-training stage primarily focuses on aligning multiple modalities with our \method LLM, thereby enabling it to capture multimodal concept representations and develop cross-modal perception capabilities.

\textbf{Stage 1. Encoder Alignment.}
This phase leverages image-text pairs, OCR data, and audio-text pairs for training, achieving alignment between the visual/audio encoders and \method LLM. Moreover, by concatenating multiple image-text pairs into a single interleaved sequence, we enhance in-context understanding capabilities and attain a $1.5\times$ acceleration in training efficiency.

\textbf{Stage 2. Image-Text Knowledge Enhancement.}
This stage concentrates on training with high-quality image-text pair data (selected from stage 1) and OCR data, with a specific focus on enhancing image-text fine-grained comprehension capabilities. This, in turn, facilitates improved understanding of interleaved image-text and video understanding tasks in stage 3. Furthermore, language-only pure text data is incorporated to prevent the degradation of \method LLM's language understanding capabilities.



\textbf{Stage 3. MultiModal Joint Training.}
This stage integrates omni-modality knowledge in a single stage, thereby facilitating comprehensive modality alignment and unified representation learning. In this stage, we incorporate high-quality image-text pairs, video-text pairs, interleaved image-text sequences, audio-text pairs, and language-only data for end-to-end multimodal pre-training.
To balance convergence rates across different modalities, a step balance strategy is employed in this stage, which will be introduced in \cref{subsubsec-Step Balancing Strategy}.


\subsubsubsection{\textbf{Instruction Tuning}}

Instruction tuning aims to make models better understand the instructions from users and fulfill the demanded tasks.

\textbf{Stage 1. Image-Text Instruction Tuning}.
This stage concentrates on enhancing the model's instruction-following ability for image modality, particularly in specialized image-related tasks, such as science, OCR, documents, and charts, which were not adequately learned during pre-training.

\textbf{Stage 2. Visual Instruction Tuning}.
This stage aims to enhance the modell's comprehensive capability on visual modality, including the capability on image-text, video-text, and interleaved image-text understanding.

\textbf{Stage 3. Omni-Modality Instruction Tuning}.
This stage further integrates the audio modality and generation tasks, enabling the model to follow instructions on mixed multi-modal sequences. Our study reveals that coordinating the training progress of diverse modalities and tasks is challenging, as the model's optimal performance across all modalities is hindered by inconsistent data volumes and convergence speeds among tasks. We propose a dynamic adaptive balance strategy to address this issue, which will be introduced in \cref{subsubsec-Dynamic Adaptive Balance}.

\subsubsubsection{\textbf{Alignment Tuning}}

This phase focuses on refining the quality and stylistic coherence of chat interactions, ensuring the model's proficiency is maintained across all modalities.  The instruction tuning stage equips the model with general multimodal conversational abilities. However, the model's responses often suffer from limitations, including brevity, lack of fluency, irrelevance, inappropriate formatting, and hallucinations, which can compromise the user experience. To mitigate these limitations and further enhance the chat experience, a preference alignment tuning stage is introduced following the instruction tuning stage. This stage employs a unified training strategy that integrates DPO~\cite{rafailov2024directpreferenceoptimizationlanguage} and instruction tuning, as defined by the equation:
\begin{equation}\label{alignment_tuning}
Loss_{at}(x) = L_{dpo}(x_{chosen}, x_{rejected})+\lambda*L_{it}(x_{chosen}, x_{it}),
\end{equation}
where $L_{dpo}$ and $L_{it}$ denotes the DPO loss and instruction tuning loss respectively. $x_{chosen}$ and $x_{rejected}$ represent the chosen samples and rejected samples from the preference dataset, and $x_{it}$ denotes the samples from the instruction dataset. In practice, we empirically set $\lambda$ to 0.3.  Additionally, we employ Low-Rank Adaptation (LoRA) \cite{hu2021loralowrankadaptationlarge} to update 5.0\% of the LLM's backbone weights, thereby preventing catastrophic forgetting.

\begin{table*}[t]
\centering
\caption{\textbf{Detailed training hyperparameters and configurations for \method.}
The model configurations are meticulously tuned to achieve consistent performance across various modalities and tasks.
Note that T, I, V, and A in the modalities row denote textual, image, video (and interleaved image-text), and audio, respectively.
U and G in the task row denote understanding and generation tasks, respectively.
LR denotes the learning rate, and AT represents the alignment tuning stage.
}
{\fontsize{8}{10}\selectfont
\renewcommand{\arraystretch}{1.0}
{
\setlength\tabcolsep{5pt}
\begin{tabular}{l|ccc|ccc|c}
\toprule
\multirow{2}{*}{Settings}         & \multicolumn{3}{c|}{Pre-training}   & \multicolumn{3}{c|}{Instruction Tuning} & AT \\
                                  & Stage 1    & Stage 2    & Stage 3    & Stage 1          & Stage 2           & Stage 3          &           \\
\hline
Data                           & \makecell{PT-Stage-\\1-Data}  & \makecell{PT-Stage-\\2-Data}    & \makecell{PT-Stage-\\3-Data}  & \makecell{IT-Stage-\\1-Data}        & \makecell{IT-Stage-\\2-Data}         & \makecell{IT-Stage-\\3-Data}        & \makecell{AT\\-Data}        \\
Modalities & I\&A        & T\&I      & All              & T\&I        & T\&I\&V         &  All  &  All      \\
Tasks  & U        & U      & U              & U        & U         &  U\&G  &  U\&G      \\
Trainable                         & Connectors        & \makecell{Vision Encoder\\+Connector}      & Full Model & Full Model              & Full Model        & \makecell{w/o\\Encoders}              & \makecell{w/o\\Encoders}       \\
LR                     & 2e-5       & 1e-5         & 2e-5       & 2e-5             & 2e-5              & 1e-5             & 5e-6             \\
LR of Encoders                     & --       & 1e-6         & 2e-6       & 2e-6             & 2e-6              & --             & --             \\
Weight Decay                      & 0.05       & 0.05         & 0.05       & 0.05             & 0.05              & 0.05             & 0.05             \\
Training Epochs                   & --         & --           & --          & 1               & 2                 & 1               & 2                \\
Max Image Tokens              & 320         & 320           & 320         & 1280               & 2560                & 2560               & 2560               \\
Max Video Frames              & --         & --           & 8         & --               & 16                & 128               & 128               \\
\bottomrule
\end{tabular}
}
}
\label{tab:app_train_hyperparameter}
\end{table*}

%% file: sections/3.approach.tex
\section{MultiModal Multi-Task Balanced Strategy}\label{sec:approach}

\subsection{Step Balance Strategy }\label{subsubsec-Step Balancing Strategy}
During the multimodal joint training stage of pre-training, two primary challenges arise: balancing of \textit{data samples} and \textit{loss weights}. On the one hand, significant disparities in data quantity exist, which hinders the model's performance on modalities with limited data. On the other hand, since losses from different modalities are not on the same scale, the training direction will be biased towards the modality task with a larger loss, leading to suboptimal convergence. To address these challenges, we propose a Step Balance strategy that combines data sample balance and loss weight balance to tackle these challenges simultaneously.

\textbf{Data Sample Balance}.
Let $\{\mathcal{D}_1, \mathcal{D}_2, ..., \mathcal{D}_M\}$ represent a collection of  $M$ different modalities of training data. Let $\mathcal{L}_i$ denote the loss function associated with the $i$-th modality.  In this pre-training stage, we explore different methods for updating the model, focusing on their effectiveness in balancing multimodal abilities. Notably, these methods are compared under the constraint that each mini-batch exclusively contains data from a single modality, ensuring a balanced and efficient training process. Three primary methods are explored:

Random Sample: A mini-batch is randomly drawn from the entire dataset. The sampling probability of each modality is proportional to its data volume:
\begin{equation}\label{random_sample}
    \theta_{t+1} = \theta_t - \eta \nabla \mathcal{L}_i(\mathcal{B}_i),
\end{equation}
where $i$ is the index of the randomly selected modality, $\mathcal{B}_i$ represents a mini-batch from modality $i$,  $\theta_t$ and $\eta$ represents the model weights at time $t$ and the learning rate, respectively.

Round-robin: We alternate mini-batches of each dataset, updating the parameters on each mini-batch. This strategy ensures equal iteration steps across modalities and facilitates sufficient training for each modality.
\begin{equation}\label{round_robin}
    \theta_{t+1} = \theta_t - \eta \nabla \mathcal{L}{i_t}(\mathcal{B}{i_t}),
\end{equation}
where  $i_t = (t \bmod M) + 1$ denotes the modality selected at time step $t$.

Accumulation: We alternate batches of each data type and compute gradients on each batch. These gradients are then weighted and summed, then are used to update the parameters.
In other words, after forward-propagating one batch from each modality, we perform a single consolidated weight update incorporating information from all modalities.
\begin{equation}\label{accumulation}
    \theta_{t+1} = \theta_t - \eta \sum_{i=1}^M \nabla \mathcal{L}_i(\mathcal{B}_i).
\end{equation}

As demonstrated by our ablation studies in \cref{{subsubsec:step_balance_ablaton}}, the accumulation method consistently exhibits superior performance compared to the other two methods, which can be attributed to its improved gradient stability and adequate training for each modality.

\textbf{Loss Weight Balance}.
A simple yet effective method is employed to determine modality-specific loss weights. The proposed method involves the following steps:
1) Train the model on a small subset of data $\mathcal{D}_i^{sub} \subset \mathcal{D}_i$ until convergence;
2) Record the converged loss value $L_i^*$;
3) Calculate the normalization weight $w_i$ using Equation \ref{eq_loss_weight}:
\begin{equation}\label{eq_loss_weight}
    w_i = \alpha\frac{1/L_i^*}{\sum_{j=1}^M 1/L_j^*}.
\end{equation}

The normalization weights are subsequently applied to the parameter gradient update, as formulated in Equation \eqref{random_sample},\eqref{round_robin},\eqref{accumulation}. In practice, for data sample balance with accumulation strategy, we set $\alpha$ to 10 and update the model parameters as:
\begin{equation}
    \theta_{t+1} = \theta_t - \eta \sum_{i=1}^M w_i \nabla \mathcal{L}_i(\mathcal{B}_i)
\end{equation}

The step balance strategy is formally presented in Algorithm \ref{alg:step_balance}, which outlines the complete procedure.




\begin{algorithm}[t]
\caption{Step Balance Training Strategy during Pre-training}
\label{alg:step_balance}
\begin{algorithmic}[1]
\State \textbf{Input:} Datasets $\{\mathcal{D}_1, ..., \mathcal{D}_M\}$
\State \textbf{Output:} Trained model parameters $\theta$
\State // small-scale datasets training to determine weights
\For{each modality $i$}
    \State Train on $\mathcal{D}_i^{sub}$ until convergence
    \State Record $L_i^*$
    \State Calculate $w_i$ using \cref{eq_loss_weight}
\EndFor
\State // Main training phase
\While{not converged}
    \For{each modality $i$}
        \State Sample batch $\mathcal{B}_i$ from $\mathcal{D}_i$
        \State Compute $\nabla_i = w_i\nabla\mathcal{L}_i(\mathcal{B}_i)$
    \EndFor
    \State Update $\theta$ using $\sum_{i=1}^M \nabla_i$
\EndWhile
\end{algorithmic}
\end{algorithm}

\subsection{\textbf{Dynamic Adaptive Balance Strategy}}\label{subsubsec-Dynamic Adaptive Balance}

\begin{figure}[t]
    \centering
    \includegraphics[width=0.8\linewidth]{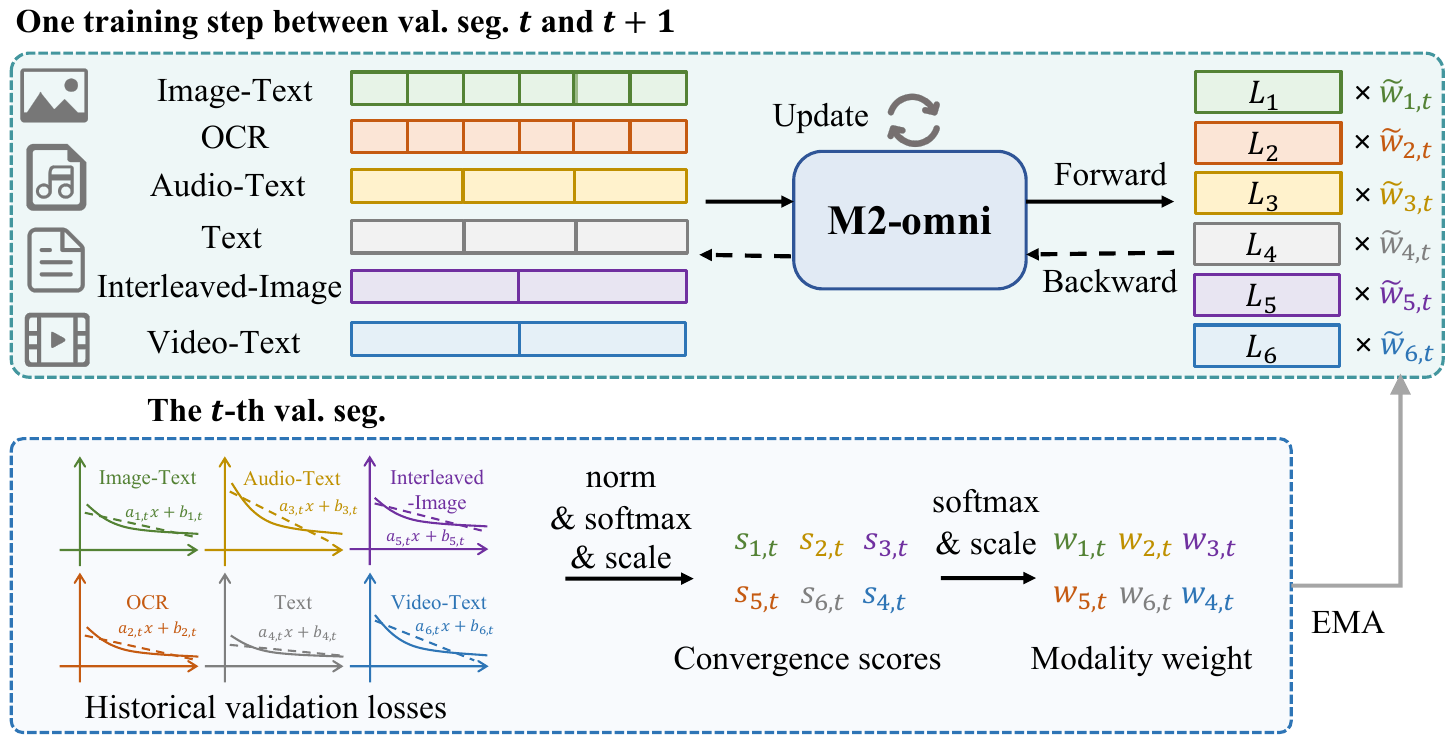}
    \caption{
    \textbf{Illustration of the dynamic adaptive balance strategy used during the Omni-Modality Instruction Tuning stage.}
    $\widetilde{w}_{i, t}$ refers to the loss weight allocated to the $i$-th modality at the $t$-th validation segment.
    }
    \label{fig-multi_modal_balance}
\end{figure}

In the Omni-Modality Instruction Tuning stage, we employ a dynamically adaptive balance strategy to regulate the convergence rates across modalities. Specifically, we treat each modality as a distinct training task, and leverage a multi-task learning (MTL)~\cite{crawshaw2020multi, liu2019loss} principle to balance the training progress of each modality (task): \textit{modalities exhibiting gentler convergence slopes received reduced weights to prevent overfitting, whereas those displaying sharper slopes were assigned higher weights to facilitate enhanced learning.}  In contrast to MTL methods~\cite{liu2024mftcoder, gong2024coba} that involve alternating training and validation cycles across entire datasets, our approach incorporates periodic validation segments at fixed intervals during training. These segments utilize a compact, predetermined validation subset to compute modality-specific validation losses, which enables us to track each modality's training progress via its validation loss and convergence slope within a historical window
This approach enables dynamic adjustment of modality weights with minimal computational overhead, thereby enhancing performance across all modalities in the context of omni-modal learning. Here, we provide a detailed introduction to our dynamically adaptive balance strategy.
The schematic diagram is shown in \cref{fig-multi_modal_balance}.

\textbf{Data Partition}. We randomly split a validation subset from the training data for each modality, comprising $\sum_{i=1}^{M}S_i*B_i$ samples. Here, $S_i$ denotes the number of validation steps per validation segment for the $i$-th modality, and $B_i$ represents the validation batch size for that modality, $M$ represents the number of modalities. The data in the validation subset is excluded from the model training process.

\textbf{Convergence Slope Calculation}. Different modalities often exhibit varying degrees of training difficulty, resulting in distinct numerical ranges of losses. To ensure fair weight allocation across modalities, we calculate the convergence slopes from the normalized validation loss:
\begin{equation}
    \widetilde{L}^{val}_{i, t} = \frac{L^{val}_{i, t} - \min\{\min\{L^{val}_{i, j}\}_{j=t-H+1}^{t}, \epsilon\}}{\max\{L^{val}_{i, j}\}_{j=t-H+1}^{t} - \min\{\min\{L^{val}_{i, j}\}_{j=t-H+1}^{t}, \epsilon\}},
\end{equation}
where $L^{val}_{i, t}$ represents the validation loss of the $i$-th modality at the $t$-th validation segment, $H$ represents the historical window size, and $\epsilon$ is set to $1e^{-6}$ to prevent division by zero. Subsequently, we employ a linear regression model of the form $a_{i, t}x+b_{i, t}$ to fit the validation loss within the historical window, where the slope coefficient $a_{i, t}$ represents the current convergence rate of the modality.

\textbf{Weight Allocation Adjustment}. To ensure a balanced initialization and mitigate potential inaccuracies in the initial convergence trajectories, we set all modality-specific loss weight $\widetilde{w}_{i, 0}$ to $1$ and maintain these fixed values throughout the first $H$ validation segments. For the $t$-th validation segment (where $t > H$), we first compute the normalized slope $\widetilde{a}_{i, t}$ and the convergence score $s_{i, t}$ for the $i$-th modality as follows:
\begin{equation}
    \widetilde{a}_{i, t} = \frac{M * a_{i, t}}{\sum_{j=1}^{M}|a_{j, t}|},
    \quad s_{i, t} = \text{softmax}\left(\widetilde{a}_{i, t}\right) * \left(-1 * \widetilde{a}_{i, t}\right),
\end{equation}
where the softmax operation is performed across the modality dimension. Next, we calculate the modal weight allocation $w_{i, t}$ of the current segment as follows:
\begin{equation}
    w_{i, t} = M * \text{softmax}\left(f * s_{i, t}\right)
\end{equation}
where $f$ is a scaling factor that regulates the probability distribution of the weights. Multiplying by $M$ ensures that the sum of the weights across all modalities equals $M$. To mitigate abrupt fluctuations caused by single-step weight updates and to enhance the stability of model training, we utilize an exponential moving average (EMA) mechanism to smoothly adjust the training weights for each modality as follows:
\begin{equation}
    \widetilde{w}_{i, t} = \alpha * \widetilde{w}_{i, t-1} + (1 - \alpha) * w_{i, t}
\end{equation}
where the smoothing factor $\alpha$ is set to $0.9$. The adjusted modality-specific loss weight $\widetilde{w}_{i, t}$ is used for all training steps until the next validation segment.

Through this dynamic adaptive balance strategy during instruction tuning, we observed that all modalities can achieve their superior performance compared to the corresponding single-modality model counterpart.

\subsection{Language Proficiency Maintenance Strategies}\label{subsubsec-Language Maintenance}
An omni-MLLM should maintain robust linguistic capabilities, as language is also one of the modalities it supports. During both the pre-training and post-training stages of our omni-MLLM, we discovered that using solely multimodal data to unfreeze our M2-omni LLM leads to a significant decline in linguistic abilities, highlighting the importance of incorporating pure-text data sources. We attribute this phenomenon to the fact that the text accompanying multimodal data often lacks the diversity and complexity of pure text data, which can lead to a biased model.  To mitigate this issue, we adopted a straightforward and efficient approach by incorporating a carefully controlled proportion of pure text data during the training whenever unfreezing the M2-omni LLM. Through extensive experimentation, we found that by carefully controlling the proportion of pure text data, specifically around 25\%, we can effectively prevent the deterioration of multimodal capabilities while maintaining robust linguistic capabilities.


\section{Data configurations}\label{subsec:app_data}

\begin{figure}[t]
    \centering
    \begin{subfigure}{0.45\textwidth}
        \centering
        \includegraphics[width=\textwidth]{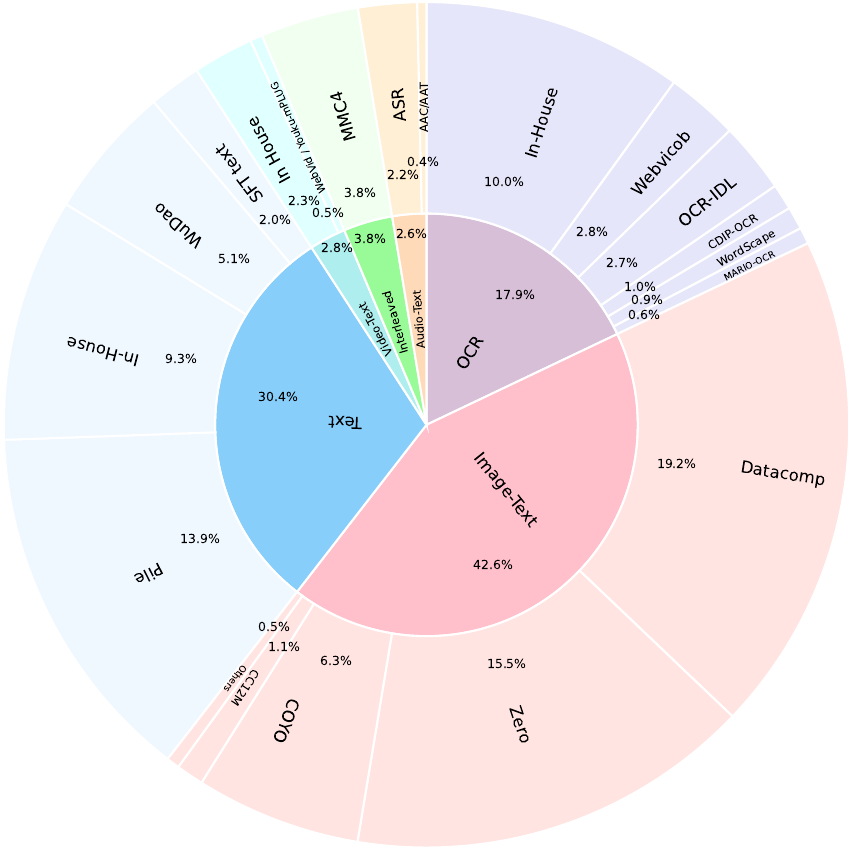}
        \vspace{-8pt}
        \caption{PT-Stag-3-Data}
        \label{fig:PT-Stage-3-Data}
    \end{subfigure}
    \begin{subfigure}{0.45\textwidth}
        \centering
        \includegraphics[width=\textwidth]{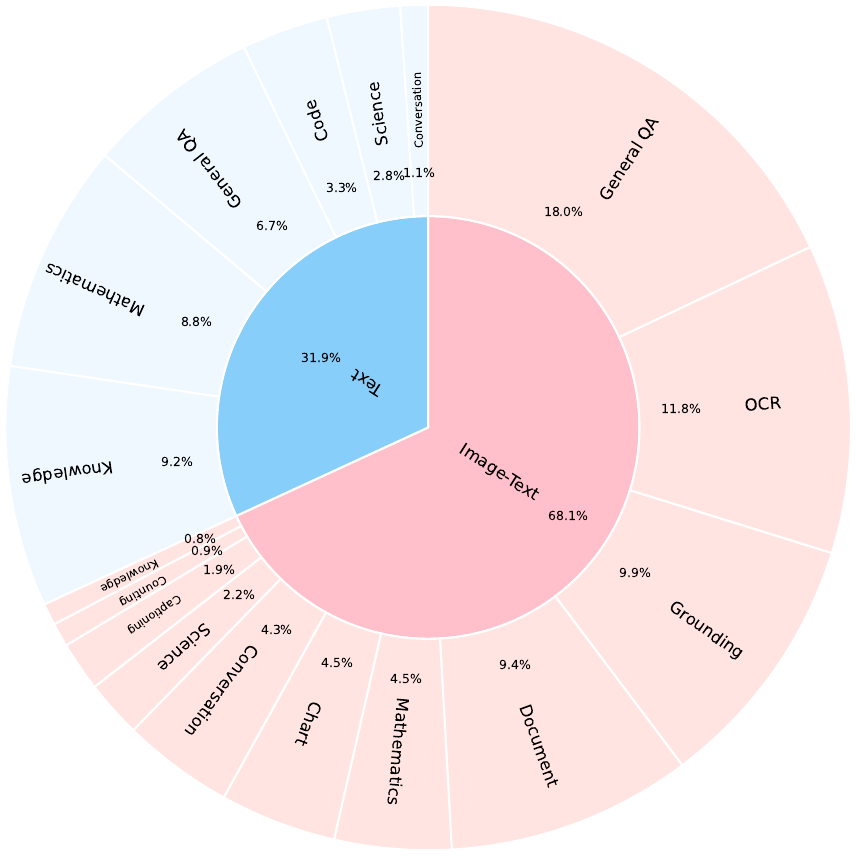}
        \vspace{-8pt}
        \caption{IT-Stage-1-Data}
        \label{fig:IT-Stage-1-Data}
    \end{subfigure}
    \begin{subfigure}{0.45\textwidth}
        \centering
        \includegraphics[width=\textwidth]{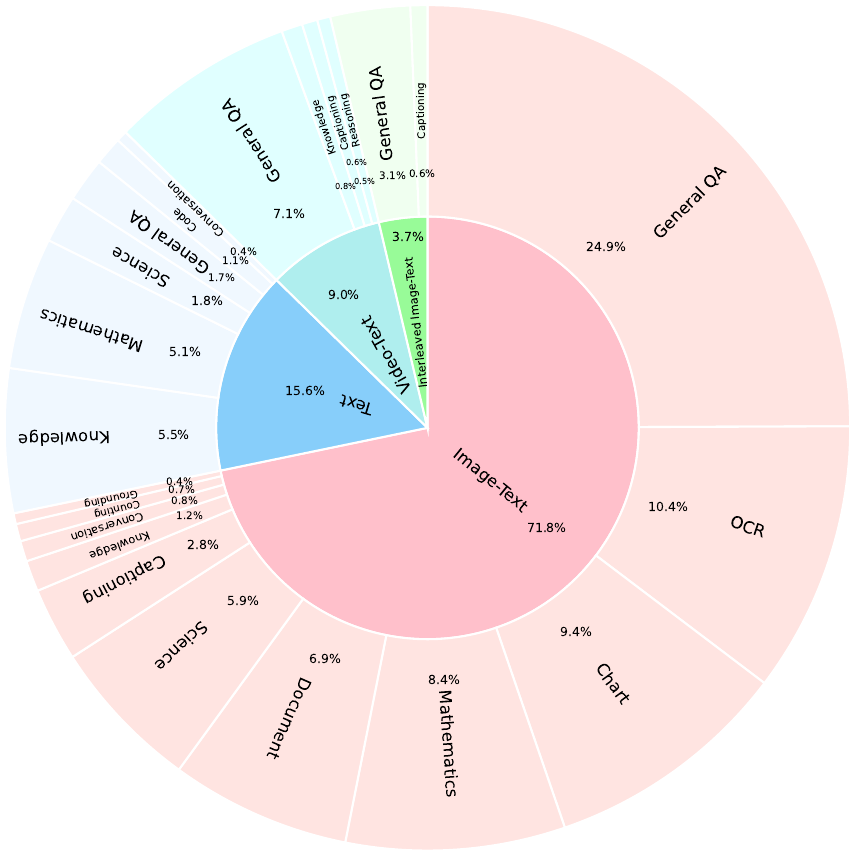}
        \vspace{-8pt}
        \caption{IT-Stage-2-Data}
        \label{fig:IT-Stage-2-Data}
    \end{subfigure}
    \begin{subfigure}{0.45\textwidth}
        \centering
        \includegraphics[width=\textwidth]{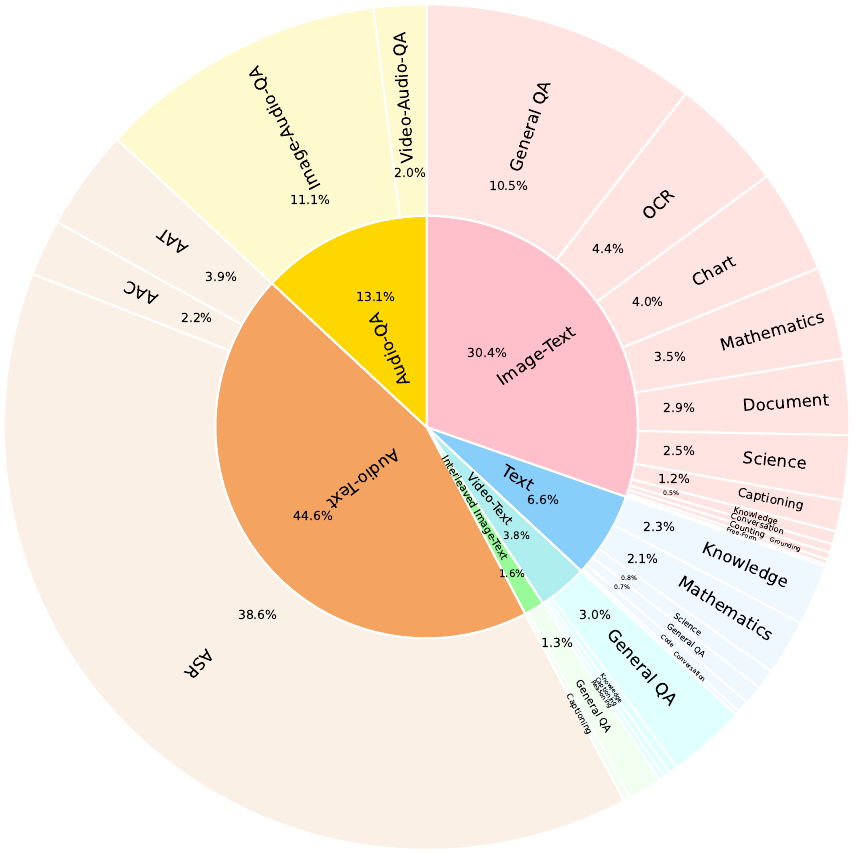}
        \vspace{-8pt}
        \caption{IT-Stage-3-Data}
        \label{fig:IT-Stage-3-Data}
    \end{subfigure}
    \caption{
    \textbf{Overview of the data configurations during pre-training and instruction tuning.}
    The numbers of image-text, OCR, text, audio-text, video-text, and interleaved image-text data are 0.33B, 0.14B, 0.15B, 25.4M, 27.3M, and 36.9M in PT-Stag-3-Data, respectively.
    The numbers of image-text and text data are 26.2M and 12.7M in IT-Stage-1-Data, respectively.
    The numbers of image-text, text, video-text, and interleaved image-text data are 17.2M, 3.7M, 2.2M, and 0.9M in IT-Stage-2-Data, respectively.
    And the numbers of image-text, text, video-text, interleaved image-text, audio-text, and audio-QA data are 17.2M, 3.7M, 2.2M, 0.9M, 25.4M, and 7.5M in IT-Stage-3-Data, respectively.
    }
    \label{fig:data_configurations}
\end{figure}

\subsection{Pre-training Data}\label{subsubsec:app_pre_training_data}
To align with the objectives of the pre-training phase, pre-training data is constructed adhering to two criteria: (1) aligning different modalities and (2) facilitating concept learning of world knowledge. Consequently, the resulting pre-training dataset exhibits diversity, aggregating multi-source data across various modalities.


[Image-Text] We primarily leverage a large-scale dataset comprising weakly labeled, web-crawled image-text pairs sourced from publicly accessible repositories, including~\cite{schuhmann2022laion,sharma2018conceptual,changpinyo2021conceptual,xiezero,byeon2022coyo,gadre2024datacomp}. To focus on high-quality knowledge learning during the Image-Text Knowledge Enhancement pretraining stage, we incorporate two comprehensive caption datasets, namely~\cite{li2024densefusion,deitke2024molmo}. The aggregated dataset consists of approximately 2 billion image-text pairs.


[OCR] To enhance performance in text-oriented tasks, we incorporate the OCR caption task. We select a range of English OCR datasets, specifically~\cite{soboroff2022complex,biten2022ocr,kim2023web,weber2023wordscape,chen2024textdiffuser}. In addition, we compile a large-scale collection of in-house Chinese OCR datasets, encompassing diverse samples from Chinese documents to scene texts.


[Audio-Text] Audio-text pairs are included for audio alignment. The dataset comprises a total of 30 million audio-text pairs, categorized into three task types: Automatic Speech Recognition (ASR)~\cite{Libriheavy, GigaSpeech, Librispeech, CoVoST2, SPGISpeech, TED_LIUM, SlideSpeech, CoVoST2, AISHELL1, WenetSpeech, KeSpeech, AliMeeting, MagicData_RAMC, Primewords_100h, FreeST, TAL, aidatatang_200zh}, Automatic Audio Caption (AAC)~\cite{WavCaps, Clotho, AudioCaps, MACS}, and Automatic Audio Tagging (AAT)~\cite{AudioSet, VGGSound}. These categories are aggregated from a diverse range of publicly accessible datasets.

[Interleaved Image-Text] Interleaved data naturally contains multiple images and accompanying text which are often interrelated, it has been demonstrated in~\cite{sun2024generative,mckinzie2024mm1} to enhance the model's few-shot capabilities in image-text tasks. The public interleaved image-text dataset Multimodal-C4 (MMC4)~\cite{zhu2024multimodal} is included.

[Video-Text] We sample video data from public datasets WebVid-10M~\cite{bain2021frozen} and Youku-mPLUG~\cite{xu2023youku} to support bilingual video-text comprehension. Besides, we collected a large-scale in-house HD video dataset.

[Text] We utilize Pile~\cite{gao2020pile} and Wudao~\cite{yuan2021wudaocorpora} datasets and supplement them with in-house text-only datasets to maintain our M2-omni LLM's linguistic abilities during pre-training.


The data configuration for each pre-training stage is carefully designed to align with the corresponding learning objectives, as illustrated in \cref{subsec:app_training}.

\textbf{PT-Stage-1-Data}. This stage utilizes coarse-grained image-text pairs, OCR data, and audio-text pairs for training, achieving alignment between visual/audio encoders and the M2-omni LLM. This stage involves a total of 2.17 billion samples.

\textbf{PT-Stage-2-Data}. We select 0.33B high-quality image-text pairs and 0.14B high-quality OCR pairs from PT-Stage-1-Data. Additionally, all detailed caption data are incorporated to enhance the model's fine-grained understanding capabilities. Furthermore, 0.15B text-only data samples, mixed with image-text instruction pairs, are included to maintain language understanding capabilities in this stage.

\textbf{PT-Stage-3-Data}.
This stage incorporates all the data from PT-Stage-2-Data, supplemented by three additional data types: audio-text, video-text, and interleaved image-text data. The data volumes of these three additional data types are determined through experiments to be 25.4M, 27.3M, and 36.9M, respectively. An overview of the data configuration of PT-Stage-3-Data is shown in \cref{fig:data_configurations}(a).

More details of publicly accessible data used in pre-training stage  can be found in \cref{tab:appendix_pretrain_data}.

\subsection{Instruction Tuning Data}\label{subsubsec:app_sft_data}


We collect both open-source and in-house multimodal instruction tuning data for training.

We first gather a comprehensive open-source dataset encompassing various multimodal instruction tuning data types, including image-text, video-text, audio-text pairs, and interleaved image-text data. High-quality language-only text instruction data are also collected to mitigate the catastrophic forgetting of the M2-omni LLM and preserve its linguistic abilities.

A primary challenge associated with the collected open-source data is the severe imbalance in data distribution.  For instance, categories like science and mathematics exhibit a significant scarcity of data, which necessitates targeted data construction efforts. To address the data imbalance, we develop a data construction pipeline that generates high-quality instruction tuning data for long-tail categories. Specifically, we establish a taxonomy of academic disciplines and analyze the data distribution to identify long-tail categories that require supplementation. Subsequently, we leverage GPT-4 ~\cite{openai2023gpt4}  to generate candidate topics for the targeted field and retrieve relevant images from web sources and our in-house data repositories. Furthermore, we employ GPT-4V~\cite{GPT4VisionSystemCard} to generate question-answer pairs related to the image, based on the candidate topics.  By leveraging this data construction engine, we generate a substantial volume of high-quality instruction tuning data.

The data configurations of each instruction tuning stage are as follows.

\textbf{IT-Stage-1-Data}.
This stage encompasses a diverse range of tasks, comprising both image-text and pure text data, as illustrated in Figure \ref{fig:data_configurations}(b).  For image-text data, we strive to ensure coverage of diverse data types while maintaining a balanced distribution. We determine an optimal ratio of approximately 30\% for text data through empirical evaluation, which helps maintain the M2-omni LLM conversational capabilities. A detailed list of the open-source data utilized in this stage is provided in \cref{tab:appendix_sft_s1_it_1,tab:appendix_sft_s1_it_2,tab:appendix_sft_s1_nlp}.



\textbf{IT-Stage-2-Data}.
In this stage, we further augment the dataset by incorporating video-text and interleaved image-text data, as illustrated in Figure \ref{fig:data_configurations}(c). Detailed lists of open-source data for image-text, video-text, and interleaved image-text are provided in Tables \ref{tab:appendix_sft_s2_it_1} and \ref{tab:appendix_sft_s2_it_2}, Table \ref{tab:video_text_stage2}, and Table \ref{tab:multi_image_stage2}, respectively.



\textbf{IT-Stage-3-Data}.
Building upon IT-Stage-2-Data, this stage introduces three new modalities: audio-text data, audio-QA data, and free-form image generation data. We create audio-QA data by converting the question text from image-text and video-text data into audio, enabling the model to comprehend images or videos based on textual or auditory instructions. In this Omni-Modality Instruction Tuning stage, we leverage text and interleaved image-text data from the previous stage, while selectively incorporating high-quality video-text data to further boost the model's video comprehension capabilities.

The detailed open-source data for video-text, audio-text, and audio-QA are provided in Tables \ref{tab:video_text_stage3}, \ref{tab:appendix_audio_text}, and \ref{tab:appendix_sft_s3_audioqa}, respectively.

\subsection{Alignment Tuning Data} \label{subsubsec:app_post_training_data}
As shown in Equation \eqref{alignment_tuning}, this stage leverages data from two primary sources: preference data and instruction tuning data. Both datasets cover a range of tasks, including multi-turn conversations, VQA, chat, charts, math, OCR, and others.

A semi-automated approach is employed to construct the preference dataset.  We first collect an in-house multimodal prompt dataset generated by actual users. Subsequently, we employ both the M2-omni model, fine-tuned through the SFT phase,  and GPT-4o to generate corresponding responses for each collected multimodal prompt. Human annotators then identify the higher-quality response from the two generated responses and construct response preference pairs. We evaluate the response quality based on four dimensions: relevance, fluency, content richness, and format rationality. More evaluation details can be found in \ref{sec:human_evaluation}.

For the instruction tuning data, we sample from each modality within the Omni-Modality Instruction Tuning stage,  and maintain a 1:1 ratio to preference data, as informed by empirical studies.



%% file: sections/4.experiment.tex
\section{Experiments}\label{sec:exp}

\begin{table}[t]
\centering
\caption{\textbf{Quantitative results on OpenCompass~\cite{2023opencompass} multimodal leaderboard.}
$^{\ddag}$ denotes closed-source models. Hall denotes HallusionBench.
}
\label{tab:exp_it_oc}
\setlength{\tabcolsep}{1pt}
\begin{tabular}{l|c|c|cccccccc}
\toprule
Models   & Params & Avg. & MM- & MM- & MM- & Math- & Hall & AI2D  & OCR- & MMVet \\
   &  &  & Bench & Star & MU & Vista &  &  & Bench & \\
\midrule
Step-1o$^{\ddag}$   & N/A   & \textbf{77.7}  & 87.3  & 69.3  & 69.9 & 74.7  & 55.8 & 89.1 & 926 & \textbf{82.8}  \\
SenseNova$^{\ddag}$  & N/A   & 77.4  & 85.7  & \textbf{72.7}  & 69.6 & \textbf{78.4}  & 57.4 & 87.8 & 894 & 78.2  \\
InternVL2.5-78B-MPO~\cite{wang2024mpo}  & 78B  & 77.0   & 87.7  & 72.1  & 68.2  & 76.6  & 58.1  & 89.2 & 909 & 73.5  \\
Qwen2.5-VL-72B~\cite{bai2025qwen25vltechnicalreport}   & 73.4B  & 76.2  & \textbf{87.8}  & 71.1  & 67.9  & 70.8  & 58.8  & 88.2  & 881  & 76.7  \\
TeleMM$^{\ddag}$   & N/A   & 75.9  & 79.9 & 70.8 & 66.6 & 75.7  & \textbf{60.6}  & 88.5 & 891 & 75.7  \\
InternVL2.5-38B-MPO~\cite{wang2024mpo}  & 38B  & 75.3  & 85.4  & 70.1 & 63.8 & 73.6 & 59.7 & 87.9 & 894 & 72.6  \\
InternVL2.5-78B~\cite{chen2024expanding}  & 78B  & 75.2 & 87.5  & 69.5 & 70 & 71.4 & 57.4 & 89.1 & 853 & 71.8   \\
Qwen2-VL-72B~\cite{qwen2-vl_2024}   & 73.4B  & 74.8  & 85.9  & 68.6  & 64.3  & 69.7  & 58.7  & 88.3  & 888  & 73.9  \\
InternVL2.5-38B~\cite{chen2024expanding}  & 38B  & 73.5  & 85.4  & 68.5  & 64.6  & 72.4  & 57.9  & 87.6  & 841  & 67.2  \\
JT-VL-Chat-V3.0$^{\ddag}$  & N/A   & 73.4  & 81.7  & 67.5  & 59.3  & 71.9  & 53.9  & 87.2  & \textbf{967}  & 69.2  \\
Taiyi$^{\ddag}$  & N/A   & 73.0  & 84.8  & 69  & 60.4  & 72.3  & 56.8  & \textbf{90.8}  & 820  & 67.9  \\
Step-1.5V$^{\ddag}$  & N/A   & 72.5 & 82.0  & 65.1  & 61.2  & 69.7  & 54.3  & 87.5  & 886  & 71.3  \\
Gemini-1.5-Pro-002$^{\ddag}$~\cite{geminiteam2024gemini15unlockingmultimodal}   & N/A   & 72.1 & 82.8  & 67.1  & 68.6  & 67.8  & 55.9  & 83.3  & 770  & 74.6  \\
InternVL2.5-26B-MPO~\cite{wang2024mpo}  & 26B  & 72.1  & 84.2  & 67.7  & 56.4  & 71.5  & 52.4  & 86.2  & 905  & 68.1  \\
GPT-4o-20241120$^{\ddag}$~\cite{openai2024gpt4ocard}  & NA   & 72.0   & 84.3  & 65.1  & \textbf{70.7}  & 59.9  & 56.2  & 84.9  & 806  & 74.5  \\
LLaVA-OneVision-72B~\cite{li2024llavaonevision}  & 73B  & 68.0  & 84.5  & 65.8  & 56.6  & 68.4  & 47.9  & 86.2  & 741  & 60.6  \\
NVLM-D-72B~\cite{nvlm2024}   & 79.4B  & 67.6  & 78.5  & 63.7  & 60.8  & 63.9  & 49.7  & 80.1  & 849  & 58.9  \\
Molmo-72B~\cite{deitke2024molmo}  & 73.3B  & 64.1  & 79.5  & 63.3  & 52.8  & 55.8  & 46.6  & 83.4  & 701  & 61.1  \\
\rowcolor{Gray} \textbf{\method-72B}   & 71.8B  & 75.1  & 86.3  & 70.7  & 57.6  & 73.3  & 56.4  & 87.6  & 889   & 79.8  \\
\midrule
\multicolumn{11}{l}{\textit{Models smaller than 20B}} \\
\midrule
Ola-7b~\cite{ola_2025}   & 8.88B   & \textbf{72.6}  & \textbf{84.3}  & \textbf{70.8}  & \textbf{57.0}  & 68.4  & \textbf{53.5}  & \textbf{86.1}  & 822  & \textbf{78.6}  \\
Qwen2.5-VL-7B~\cite{bai2025qwen25vltechnicalreport}   & 8.29B   & 70.4  & 82.6  & 64.1  & 56.2  & 65.8  & 56.3  & 84.1  & 877  & 66.6  \\
InternVL2.5-8B-MPO~\cite{wang2024mpo}   & 8B   & 70.3  & 82  & 65.2  & 54.8  & 67.9  & 51.7  & 84.5  & \textbf{882}  & 68.1  \\
MiniCPM-o-2.6~\cite{yao2024minicpm}   & 8.67B   & 70.2  & 80.6  & 63.3  & 50.9  & \textbf{73.3}  & 51.1  & 86.1  & 889  & 67.2  \\
Ovis1.6-Gemma2-9B~\cite{lu2024ovis}  & 10.2B  & 68.8  & 80.5  & 62.9  & 55.0  & 67.2  & 52.2  & 84.4  & 830  & 65.0  \\
InternVL2.5-8B~\cite{chen2024expanding}   & 8B   & 68.1  & 82.5  & 63.2  & 56.2  & 64.5  & 49.0  & 84.6  & 821  & 62.8  \\
POINTS1.5-Qwen2.5-7B~\cite{points1.5_2024} & 8.3B   & 67.4  & 80.7  & 61.1  & 53.8  & 66.4  & 50.0  & 81.4  & 832  & 62.2  \\
Valley-Eagle$^{\ddag}$   & 8.9B   & 67.4  & 80.7  & 60.9  & \textbf{57.0}  & 64.6  & 48.0  & 82.5  & 842  & 61.3  \\
Qwen2-VL-7B~\cite{qwen2-vl_2024}  & 8B   & 67.0  & 81.0 & 60.7 & 53.7 & 61.4  & 50.4 & 83 & 843 & 61.8 \\
DeepSeek-VL2~\cite{wu2024deepseekvl2}   & 16.1B  & 66.4  & 81.2  & 61.0  & 50.7  & 59.4  & 51.5  & 84.5  & 825  & 60.0  \\
VITA-1.5~\cite{fu2025vita}   & 8.3B   & 63.3  & 76.8  & 60.2  & 52.6  & 66.2  & 44.6  & 79.2  & 741  & 52.7  \\
Baichuan-Omni~\cite{baichuan-omni}   & 7B   & -  & 75.6  & -  & 47.3  & 51.9  & 47.8  & -  & 700  & 65.4  \\
LLaVA-OneVision-7B~\cite{li2024llavaonevision}   & 8B   & 61.2  & 76.8  & 56.7  & 46.8  & 58.5  & 47.5  & 82.8  & 697  & 50.6  \\
Molmo-7B-D~\cite{deitke2024molmo}   & 8B   & 58.9  & 70.9  & 54.4  & 48.7  & 47.3  & 47.7  & 79.6  & 694  & 53.3  \\
\rowcolor{Gray} \textbf{\method-9B}  & 8.8B   & 69.7  & 80.7  & 60.5  & 51.2  & 68.3  & 51.8  & 84.5  & 883 & 72.3 \\
\bottomrule
\end{tabular}
\end{table}

\begin{table}[t]
  \caption{\textbf{Performance comparison on video and Interleave benchmarks} compared with existing approaches. $^*$ indicates officially released checkpoints evaluated by us. Best performance is marked \textbf{bold}. }
  \label{tab: video_n_interleave}
  \centering
  \setlength{\tabcolsep}{7.5pt}
  \begin{tabular}{lccccc}
    \toprule
       & \multicolumn{2}{c}{\textbf{VideoMME}} & \multicolumn{1}{c}{\textbf{MVBench}} & \multicolumn{2}{c}{\textbf{Llava-Interleave}}\\
    \cmidrule(r){2-3} \cmidrule(r){4-4} \cmidrule(r){5-6}
    Model & w/o subs & w subs & avg & in-domain & out-domain \\
    \midrule
     MiniCPM-V-2.6~\cite{yao2024minicpm} &  60.9 &  63.6 &  - &  - &  - \\
     LLaVA-OneVision-7B~\cite{li2024llavaonevision} &  58.2 &  - &  - &  - &  - \\
     Qwen2-VL-7B~\cite{qwen2-vl_2024} &  63.3 &  69.0 &  67.0 &  49.5$^*$ &  51.0$^*$ \\
     InternVL2-8B~\cite{chen2024far} &  56.3 & 59.3 &  65.8 &  - &  - \\
     VITA-1.5~\cite{fu2025vita} &  56.1 & 58.7 &  55.4 &  - &  - \\
     Baichuan-Omni~\cite{baichuan-omni} &  58.2 & - &  60.9 &  - &  - \\
     MiniCPM-o-2.6~\cite{yao2024minicpm} & 63.0$^*$ & 65.3$^*$ & 58.1$^*$ &  43.5$^*$ &  36.8$^*$ \\
     \rowcolor{Gray} \textbf{\method-9B} &  60.4  & 65.0 &  66.3 &  59.8 &  87.8 \\
    \midrule
    VideoLLaMA2-72B~\cite{cheng2024videollama2} & 61.4 & 63.1 & 62.0 & - & - \\
    LLaVA-OneVision-72B~\cite{li2024llavaonevision} &  66.2 &  69.5 &  59.4 &  - &  - \\
    Qwen2-VL-72B~\cite{qwen2-vl_2024} &  71.2 &  77.8 &  \textbf{73.6} &  - &  - \\
    InternVL2-Llama3-76B~\cite{chen2024far} &  64.7  & 67.8 &  69.6 &  - &  - \\
    \rowcolor{Gray} \textbf{\method-72B} &  65.2  & 67.7 &  69.6 &  \textbf{63.5} &  \textbf{89.9} \\
    \midrule
    GPT-4v~\cite{GPT4VisionSystemCard} & 59.9 & 63.3 & 43.7 & 39.2 & 57.78 \\
    GPT-4o-20240513~\cite{openai2024gpt4ocard} & 71.9 & 77.2 & - & - & - \\
    Gemini-1.5-Pro~\cite{geminiteam2024gemini15unlockingmultimodal} & \textbf{75.0} & \textbf{81.3} & - & - & - \\
    \bottomrule
\end{tabular}
\end{table}

In this section, we present a comprehensive evaluation of our \method model, comprising both quantitative and qualitative analyses of its performance. Furthermore, we conduct ablation studies to analyze the contributions of several key design components to the performance of our \method model, providing insights into their distinct impacts.


\subsection{Quantitative Results}\label{subsec:exp_quantitative_results}

\subsubsection{Image-Text Understanding}
To evaluate the effectiveness of our \method in image-text understanding, we benchmark it against state-of-the-art MLLMs on the OpenCompass~\cite{2023opencompass} multimodal leaderboard, a widely recognized platform for multimodal evaluation. This leaderboard contains 8 different multimodal benchmarks, including complex VQA (MMBench~\cite{liu2025mmbench}, MMStar~\cite{chen2024we}, MMMU~\cite{yue2023mmmu}, AI2D~\cite{kembhavi2016diagram}, and MMVet~\cite{yu2024mm}), multimodal reasoning (MathVista~\cite{lu2024mathvista}), hallucination evaluation (Hallusionbench~\cite{Guan_2024_hallusionbench}), and OCR (OCRBench~\cite{Liu_2024}).
\cref{tab:exp_it_oc} shows the overall results. Our \method-72B model achieves top-tier performance on most benchmarks, surpassing closed-source models like GPT-4o and Gemini-1.5-Pro. Furthermore, our \method-9B model exhibits competitive performance among models of similar size, showcasing its robust capabilities in image-text understanding tasks.


\subsubsection{Video \& Interleaved Image-Text Understanding}

We evaluate our model's video and interleaved image-text understanding abilities on three mainstream benchmarks.

\textbf{Video-MME}~\cite{fu2024video}: Video-MME is a benchmark designed to evaluate MLLMs in full-spectrum video analysis. It encompasses a wide variety of video types across multiple domains and durations, featuring multimodal inputs such as video, subtitles, and audio. For this benchmark, testing is conducted with under 96 frames, and results are reported for both "with subtitles" and "without subtitles" settings.

\textbf{MVBench}~\cite{li2024mvbench}: MVBench serves as a video understanding benchmark aimed at thoroughly evaluating the temporal awareness of MLLMs in an open-world context. It includes 20 challenging video tasks that range from perception to cognition, which cannot be adequately addressed using a single frame. Testing for this benchmark utilizes dynamic sampling frames.

\textbf{LLaVA-Interleave Bench}~\cite{llava-next_2024}: LLaVA-Interleave Bench comprises a comprehensive suite of multi-image benchmarks collected from public datasets or generated via the GPT-4V API. It is created to assess the interleaved multi-image reasoning capabilities of MLLMs, with reported results for both "in-domain" and "out-domain" subsets.

As shown in Table~\ref{tab: video_n_interleave}, \method-9B achieves the second-best results across VideoMME and MVBench (outperformed only by Qwen2-VL-7B but requiring significantly fewer frames). However, the performance gains do not scale up to \method-72B due to limitations in the quantity of instruction-tuned video data. Moreover, both our \method-9B and \method-72B greatly surpass all other baselines in multi-image benchmarks, both in-domain and out-of-domain, highlighting their potential as strong competitors for complex tasks.

\subsubsection{Audio Understanding}

We evaluate our M2-omni model's audio understanding abilities on four mainstream benchmarks.

\textbf{Multilingual LibriSpeech (MLS)}~\cite{MLS_English}: The Multilingual LibriSpeech dataset is an extensive collection of read audiobooks sourced from Librivox, available in eight different languages. We utilize the English test set from this dataset to assess the model's speech comprehension capabilities. The latest version of this corpus comprises approximately 50,000 hours.

\textbf{Librispeech}~\cite{Librispeech}: The Librispeech corpus comprises approximately 1,000 hours of transcribed speech audio data derived from read English audiobooks. The entire dataset is categorized into three training sets (100 hours of clean, 360 hours of clean, and 500 hours of other), two validation sets (clean and other), and two test sets (clean and other). In this study, we assess our model's audio comprehension capabilities using both the clean and other testsets.

\textbf{Aishell1}~\cite{AISHELL1}:  The Aishell1 dataset comprises 178 hours of speech data, recorded by 400 speakers from various accent regions across China. It is organized into three subsets: a training set consisting of 340 speakers, a validation set with 40 speakers, and a test set featuring 20 speakers.

\textbf{AudioCaps}~\cite{AudioCaps}: AudioCaps is a comprehensive dataset featuring audio event descriptions specifically curated for the purpose of audio captioning. The sounds within this collection are derived from the AudioSet dataset. We utilize this dataset to assess the audio captioning capabilities of our \method.

The results are presented in Table~\ref{tab:exp_audio_understand}, and our \method-9B demonstrates competitive performance in speech recognition and audio captioning tasks. 
Specifically, our \method-9B is comparable to GPT-4o-Realtime~\cite{openai2024gpt4ocard}.
In addition, \method-9B significantly outperforms all other baselines on AudioCaps benchmarks, while achieving the second-best results for the MLS English, Librispeech other, Librispeech-clean and Aishell1 benchmarks.

\begin{table}[]
\centering
\caption{\textbf{Quantitative results on speech recognition and audio captioning.}
 $^*$ indicates results from \cite{yao2024minicpm}.
}
\label{tab:exp_audio_understand}
\setlength{\tabcolsep}{7pt}
\begin{tabular}{l|cccccccccc}
\toprule
Models   & MLS- & Librispeech- & Librispeech- & Aishell1 & AudioCaps \\
                & English & other & clean &  & \\
                & WER$\downarrow$ & WER$\downarrow$ & WER$\downarrow$ & WER$\downarrow$ & CIDER$\uparrow$ \\
\midrule
UIO2-L-1.1B~\cite{lu2023uio2}   & - & - & - & - & 45.7   \\
UIO2-XL-3.2B~\cite{lu2023uio2}  & - & - & - & - & 45.7   \\
UIO2-XXL-6.8B~\cite{lu2023uio2} & - & - & - & - & 48.9  \\
Whisper-large-v2~\cite{Whisper}  & \textbf{6.83} & \textbf{5.16} & 2.87 & - & - \\
Paraformer-cn~\cite{gao2022paraformer} & - & - & - & 2.12 & - \\
VITA-1.5~\cite{VITA_1.5} & - & 7.5 & 3.4 & 2.2 & - \\
Mini-Omini2~\cite{mini_omni2} & - & 9.8 & 4.8 & - & - \\
Freeze-Omini~\cite{Freeze_Omni} & - & 10.5 & 4.1 & 2.8 & - \\
MiniCPM-o-2.6~\cite{yao2024minicpm} & - & - & \textbf{1.7} & \textbf{1.6} & - \\
GPT-4o-Realtime~\cite{openai2024gpt4ocard} & - & - & 2.6$^*$ & 7.3$^*$ & - \\
\rowcolor{Gray} \textbf{\method-9B}   & 7.19 & 5.29 & 2.07 & 1.99 & \textbf{49.2} \\
\bottomrule
\end{tabular}
\end{table}

\begin{table}[t]
\centering
\caption{\textbf{Quantitative results on language benchmarks.} $^*$ indicates officially released checkpoints evaluated using the tools provided by OpenCompass~\cite{2023opencompass}.
}
\label{tab:exp_language}
\setlength{\tabcolsep}{5pt}
\begin{tabular}{cccccccc}
\hline
Tasks & MMLU & AGIEVAL & ARC-C & GPQA & MATH & HellaSwag & \begin{tabular}[c]{@{}l@{}}Avg.\\ Accuracy\end{tabular} \\ \hline
LLama3.1-8B & 69.4 & 41.2$^*$ & 83.4 & 30.4 & 51.9 & 75.1$^*$ & 58.6  \\
\rowcolor{Gray} \textbf{\method-9B} & 68.5 & 43.7 & 78.7 & 32.3 & 51.8 & 80.1 & 59.2  \\ \hline
\end{tabular}
\end{table}

\subsubsection{Audio Generation}
In this section, we also evaluated our model on the commonly-used test set: SEED-TTS test-zh. \textbf{SEED-TTS}~\cite{SEED_TTS} serves as an out-of-domain evaluation test set, comprising diverse input texts and reference speeches from various domains. We present the experimental results for \method-9B and the baseline models in Table~\ref{tab:exp_audio_generation}. As shown in Table~\ref{tab:exp_audio_generation}, our model outperforms MiniCPM-o-2.6~\cite{yao2024minicpm} in speech generation capability, achieving significant improvements in both evaluation metrics. However, our \method-9B still lags behind traditional vertical speech generation models, highlighting the need for further research and development to bridge this gap.

\subsubsection{Text-only Performance}
In this section, we assess the performance of our proposed \method-9B model and its initial counterpart, Llama3.1-8B~\cite{llama3_2024}. To evaluate the models' knowledge and examination capabilities, we employ a range of benchmarks, including AGIEVAL~\cite{zhong2023agievalhumancentricbenchmarkevaluating} and MMLU~\cite{hendrycks2021measuringmassivemultitasklanguage}. Furthermore, we utilize a diverse set of benchmarks to evaluate the models' multi-step problem-solving capabilities, including MATH~\cite{hendrycks2021measuringmathematicalproblemsolving} for mathematical derivation, HellaSwag~\cite{zellers2019hellaswagmachinereallyfinish} for commonsense reasoning in real-world contexts, ARC-C~\cite{allenai:arc} for scientific logical chains, and GPQA~\cite{rein2023gpqagraduatelevelgoogleproofqa} for critical analysis in expert-level domains. For all evaluation datasets, we adopt a generation-based assessment approach with greedy decoding.

Our experimental results, presented in \cref{tab:exp_language}, demonstrate that the performance of our proposed \method-9B model outperforms its initial counterpart, Llama3.1-8B across most evaluation datasets,   which is attributed to our multi-stage language preservation strategy and the high-quality instruction tuning data used in our training process.


\begin{table}[t]
\centering
\caption{
\textbf{Free-form dialogue generation evaluation results.}
}
\setlength{\tabcolsep}{8pt}
\begin{tabular}{c|c|c|c}
\toprule
Model & Relevance & Fluency & Informativeness\\
\midrule
TextBind~\cite{li2023textbind} & 3.85 & 4.30 & 3.25\\
\rowcolor{Gray} \textbf{\method-9B} & 4.60 & 4.80 & 3.80\\
\bottomrule
\end{tabular}
\label{tab-model_freeform_results}
\end{table}


\begin{table}[t]
  \caption{\textbf{Quantitative results on audio generation.} $^*$ indicates officially released checkpoints evaluated by us.}
  \label{tab:exp_audio_generation}
  \centering
  \setlength{\tabcolsep}{14pt}
  \begin{tabular}{lccccc}
    \toprule
       & \multicolumn{2}{c}{\textbf{SEED test-zh}}\\
    \cmidrule(r){2-3}
    Model & CER(\%)$\downarrow$ & SS$\uparrow$  \\
    \midrule

     Human & 1.26 &0.755 \\
     Vocoder Resyn. & 1.27 & 0.720 \\
     \midrule
     Seed-TTS~\cite{SEED_TTS} & 1.12 & 0.796 \\
     FireRedTTS~\cite{FireRedTTS} & 1.51 &0.635 \\
     MaskGCT~\cite{MaskGCT} & 2.27 & 0.774 \\
     E2-TTS(32 NFE)~\cite{E2_TTS} & 1.97 & 0.730 \\
     F5-TTS(32 NFE)~\cite{F5_TTS} & 1.56 & 0.741 \\
     CosyVoice~\cite{CosyVoice} &3.63 &0.723 \\
     CosyVoice2~\cite{CosyVoice2} &1.45 &0.748 \\
     CosyVoice2-S~\cite{CosyVoice2} &1.45 &0.753 \\
     CosyVoice2-S~\cite{CosyVoice2} &1.45 &0.753 \\
     \midrule
     MiniCPM-o-2.6~\cite{yao2024minicpm} &8.03$^*$ &0.474$^*$ \\
     \rowcolor{Gray} \textbf{\method-9B} &  6.36  & 0.604 \\
    \bottomrule
\end{tabular}
\end{table}

\subsubsection{User Experience Evaluation}\label{sec:human_evaluation}
\textbf{Evaluation Metric}:
Current benchmarks such as MMBench~\cite{liu2025mmbench}, MMStar~\cite{chen2024we}, and MMMU~\cite{yue2023mmmu} primarily focus on assessment through judgment-style questions. However, this assessment does not align with the users' actual interactive experience with MLLMs. To address this limitation, drawing inspiration from SuperclueV~\cite{supercluev}, we develop evaluation criteria specifically for assessing the models' performance on user experience, which contains four key dimensions: relevance, fluency, informativeness, and format rationality. \textit{Relevance} assesses the extent to which the model's responses align with both the provided prompts and the multimodal inputs.
\textit{Fluency} evaluates the naturalness, smoothness, clarity, comprehensibility, and anthropomorphic quality of the model's responses.
\textit{Informativeness} measures the extent to which the model's responses provide relevant information, knowledge, and analytical reasoning, enhancing their utility, detail, depth, and innovation.
\textit{Format rationality} examines the model's ability to adaptively generate appropriately structured and clear formats, for presenting results based on varying prompt types.


\begin{table}[t]
\centering
\caption{
\textbf{Detailed model experience evaluation standards.}
}
\setlength{\tabcolsep}{4pt}
\begin{tabular}{c|c}
\toprule
Score & Description\\
\midrule
1 & Totally unsatisfied, totally unacceptable \\
2 & Basically not satisfied, with many obvious problems \\
3 & Generally satisfied, with a few obvious problems \\
4 & Basically satisfied, minor flaws allowed \\
5 & Completely satisfied, almost perfect \\
\bottomrule
\end{tabular}
\label{tab-model_expr_standards}
\end{table}

\textbf{Evaluation Dataset}: We collect chat samples from the actual users' multi-turn interaction dialogues, which cover a variety of tasks, including visual question answering (VQA), conversational interactions, chart interpretation, mathematical problem-solving, optical character recognition (OCR), and other related tasks. GPT-4o~\cite{openai2024gpt4ocard} is instructed to follow the evaluation criteria to generate initial reference answers for these collected samples. To ensure accuracy, human annotators refine the initial responses generated by GPT-4o. This process yields an evaluation dataset with nearly 300 samples, each with a corresponding ground truth.

We utilize GPT-4o to evaluate the model's responses against the ground truth, adhering to the standards outlined in  \cref{tab-model_expr_standards}.  As shown in \cref{tab-user_experience},  our M2-omni model, after undergoing  alignment tuning,  demonstrates an average increase of 5.7\%-23.4\% in user experience performance, which is further validated by human annotations on selected cases. Meanwhile, our model's performance on the OC benchmark across other modalities remains relatively consistent, thereby demonstrating the effectiveness of our unified training strategy, which integrates DPO and instruction tuning in the alignment tuning stage.


\subsubsection{Free-Form Dialogue Generation}
To evaluate the open-world multi-turn multimodal instruction following capabilities of our model, we create a test set consisting of 50 conversations derived from realistic scenarios. We utilize \method-9B to generate arbitrarily interleaved text and images in proper conversation contexts.
For quantitative results, following our user experience evaluation metric, we employ GPT-4o to rate each conversation on a scale of 0 to 5 across three evaluation dimensions: relevance, fluency, and informativeness.
We carry out our quantitative results against recent work TextBind~\cite{li2023textbind}. As shown in \cref{tab-model_freeform_results}, \method-9B exhibits overall better understanding and generating ability of multi-turn multimodal conversations. More qualitative cases can be found in \cref{fig-IT-Freeform-Result}.

\begin{table}[t]
\centering\footnotesize
\caption{
\textbf{Detailed evaluation on user experience benchmark and OC benchmark. OC is short for the OpenCompass image-text understanding benchmark.}
}
\setlength{\tabcolsep}{3pt}
\begin{tabular}{c|c|c|c|c|c|c}
\toprule
Model & Relevance & Fluency & Informativeness & Format Rationality & Expr. Avg($\Delta$\%) & OC Avg($\Delta$)\\
\midrule
\method-9B & 4.556 & 4.036 & 2.742 & 3.573 & 3.726 & -\\
\rowcolor{Gray} \method-9B-Align & 4.893 & 4.735 & 4.118 & 4.644 & 4.598(+23.4\%) & -0.3\\
\method-72B & 4.942 & 4.689 & 3.267 & 4.265 & 4.351 & -\\
\rowcolor{Gray} \method-72B-Align & 4.946 & 4.875 & 3.961 & 4.615 & 4.598(+5.7\%) & -0.2\\
InternVL2-26B~\cite{internvl_2024} & 4.886 & 4.76 & 4.15 & 4.52 & 4.577 & -\\
GPT-4o~\cite{openai2024gpt4ocard} & 5 & 4.878 & 3.854 & 4.831 & 4.64 & -\\
\bottomrule
\end{tabular}
\label{tab-user_experience}
\end{table}

\subsection{Qualitative Results}\label{subsec:exp_qualitative_results}

In this section, we qualitatively assess the capabilities of our \method, presenting examples of each modality and different tasks.

\begin{figure}[t]
    \centering
    \includegraphics[width=0.9\linewidth]{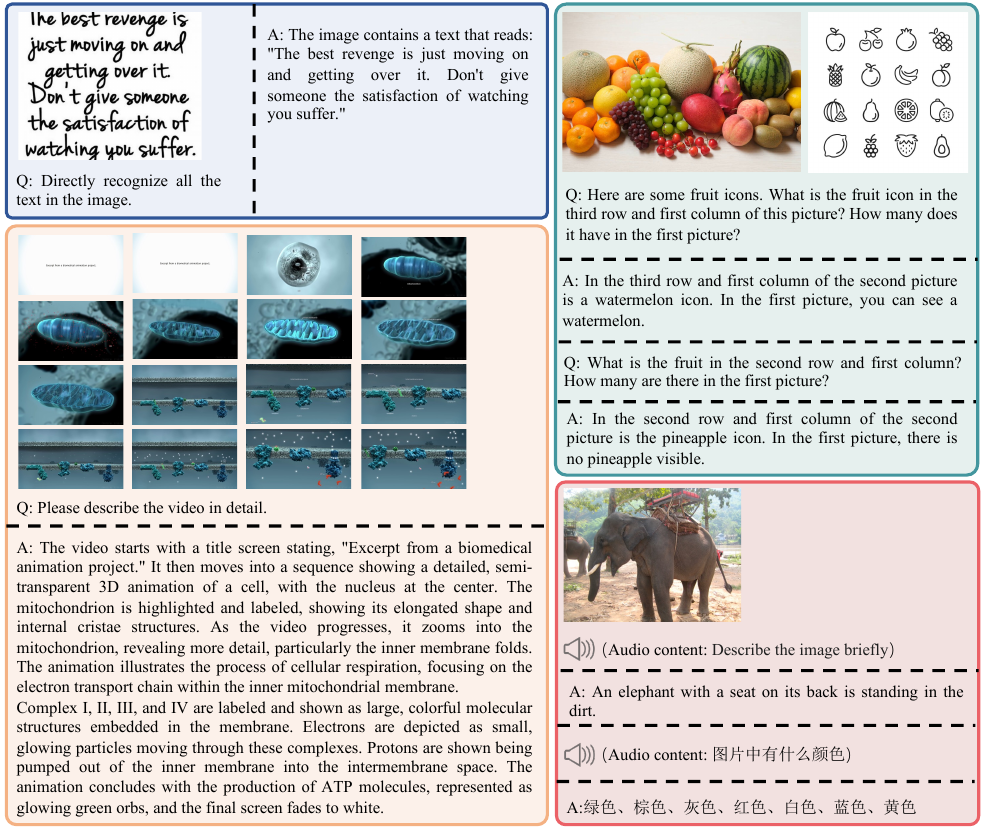}
    \caption{
    \textbf{Cases for multimodal understanding.}
    \method shows great potential to solve various multimodal problems.
    }
    \label{fig-exp_case_all}
\end{figure}

We show multimodal understanding abilities of our \method in \cref{fig-exp_case_all}. \method demonstrates promising capabilities in processing cross-modal problems, encompassing image understanding, video understanding, interleaved image-text understanding, and image-audio understanding. More examples can be found in the appendix, provided in \cref{subsec:appendix_cases}.

\cref{fig-IT-Freeform-Result} illustrates the model's ability to generate free-form dialogue, where our \method can create images based on the conversation context without explicit user input, useful for explaining ideas to users.

\begin{figure}[t]
    \centering
    \includegraphics[width=0.9\linewidth]{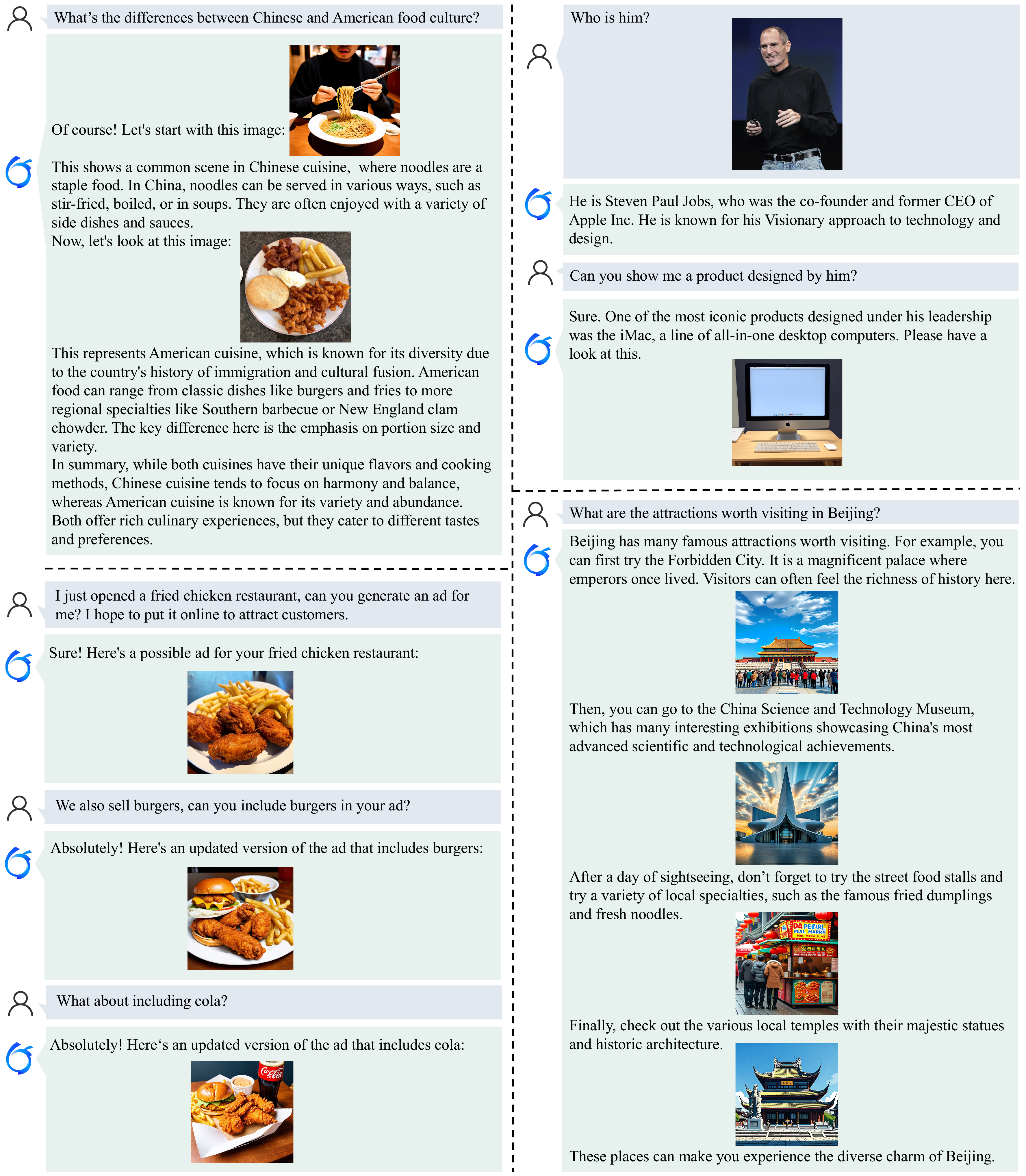}
    \caption{
    \textbf{Cases for Free-Form Dialogue Generation.}
    }
    \label{fig-IT-Freeform-Result}
\end{figure}

\subsection{Ablation Study}\label{subsec:exp_ablation}

\begin{table}[t]
\centering
\caption{\textbf{Ablation studies on step balancing strategy.} The loss weight setting [1,1,1] corresponds to the uniform weighting of the loss functions for image-text pairs, interleaved image-text, and video datasets.  * and \# represent the loss weight settings. * is obtained through experimental trials and parameter tuning. \# is obtained by normalizing the loss weights using the inverse of the loss at convergence, as described in Section \cref{subsubsec-Step Balancing Strategy}. We evaluate the few-shot performance on VQA tasks and the zero-shot performance on the captioning task of our pre-trained model.}
\label{tab:ablation_step_balance_pretrain}
\setlength{\tabcolsep}{4pt}
\begin{tabular}{c|c|ccc}
\toprule
\multicolumn{1}{l|}{Data Sample Balance} & Loss Weight Balance & \multicolumn{1}{l}{OK-VQA(4-shot)} & \multicolumn{1}{l}{VQAv2(4-shot)} & \multicolumn{1}{l}{Flickr30k(0-shot)} \\ \hline
Random Sample                        & {[}1,1,1{]}          & 40.5                             & 54.3                             & 87.0                                 \\
Round-robin                          & {[}1,1,1{]}          & 41.6                             & 54.4                             & 88.1                                 \\
Accumulation                         & {[}1,1,1{]}          & 41.7                             & 54.6                             & 88.2                                 \\
Accumulation                         & ${[}0.2,1.0,0.03{]}^{*}$   & 39.7                             & 52.5                             & 87.1                                 \\
Accumulation                         & ${[}0.45,0.36,1.09{]}^{\#}$ & \textbf{42.1}                             & \textbf{55.4}                             & \textbf{88.2}                                 \\
\bottomrule
\end{tabular}
\end{table}

In this section, we conduct ablation studies to investigate the effectiveness of our step balance strategy and dynamic adaptive balance strategy in our M2-omni model. These experiments aim to provide insights into the impact of these key components on our M2-omni’s performance.

\subsubsection{Step Balance Strategy}\label{subsubsec:step_balance_ablaton}

As described in \cref{subsubsec-Step Balancing Strategy} ,  we investigate the impact of various data sample balancing strategies and loss weight balancing schemes on the multimodal joint training stage of pre-training. We evaluate the performance of candidate strategies on two VQA benchmarks, OK-VQA~\cite{marino2019ok} and VQAv2~\cite{goyal2017making}, and assess its image captioning performance using the Flickr30k~\cite{young2014image} benchmark.

For pretrained models lacking in instruction following ability, to assess the effectiveness of our approach, we evaluate the performance of these models on VQA tasks using a few-shot approach and on image caption tasks using a zero-shot approach. \cref{tab:ablation_step_balance_pretrain} presents the results of our M2-omni pretrained models, which demonstrate the effectiveness of our step balance strategy.



\begin{table}[t]
\centering
\caption{
\textbf{Ablation results of the dynamic adaptive balance strategy}. Results for unimodal baselines are derived from the following single-modal models: \textsuperscript{$\dagger$} Image-Text Model, \textsuperscript{$\ddagger$} Video-Text Model, and \textsuperscript{
$\mathsection$} Audio-Text Model. The best result for each benchmark is \textbf{bolded}, while the best result for each model across all epochs is \underline{underlined}.
}
\setlength{\tabcolsep}{3pt}
\begin{tabular}{l|l|ccccc|cc|cc}
\toprule
Models &  & MM- & OK- & VQAv2 & Text- & GQA & MSVD- & MSRVTT & Audio & MLS- \\
& & Bench & VQA &&VQA&& QA & QA & Caps & English($\downarrow$) \\
\midrule
\multirow{3}{*}{\makecell[l]{Single-modal\\Baselines}} & ep1 & 68.0\textsuperscript{$\dagger$} & 56.4\textsuperscript{$\dagger$} & 74.8\textsuperscript{$\dagger$} & \underline{70.4}\textsuperscript{$\dagger$} & 58.4\textsuperscript{$\dagger$} & 72.3\textsuperscript{$\ddagger$} & 59.3\textsuperscript{$\ddagger$} & 29.0\textsuperscript{$\mathsection$} & 11.4\textsuperscript{$\mathsection$} \\
& ep2 & \underline{\textbf{77.8}}\textsuperscript{$\dagger$} & \underline{59.8}\textsuperscript{$\dagger$} & \underline{76.9}\textsuperscript{$\dagger$} & 69.8\textsuperscript{$\dagger$} & 60.6\textsuperscript{$\dagger$} & \underline{\textbf{76.5}}\textsuperscript{$\ddagger$} & \underline{60.1}\textsuperscript{$\ddagger$} & \underline{39.9}\textsuperscript{$\mathsection$} & 9.33\textsuperscript{$\mathsection$} \\
& ep3 & 77.3\textsuperscript{$\dagger$} & 58.0\textsuperscript{$\dagger$} & 76.8\textsuperscript{$\dagger$} & 69.1\textsuperscript{$\dagger$} & \underline{60.8}\textsuperscript{$\dagger$} & 74.4\textsuperscript{$\ddagger$} & 58.6\textsuperscript{$\ddagger$} & 39.5\textsuperscript{$\mathsection$} & \underline{8.96}\textsuperscript{$\mathsection$} \\
\midrule
\multirow{3}{*}{\makecell[l]{Mixture \\w/o MM-Bal.}}
& ep1 & 70.5 & 55.9 & 75.7 & 70.2 & 57.7 & \underline{75.1} & \underline{59.6} & 27.5 & 12.1 \\
& ep2 & \underline{75.8} & \underline{58.8} & \underline{77.0} & \underline{70.5} & \underline{\textbf{61.1}} & 73.4 & 58.5 & 33.5 & 9.45 \\
& ep3 & 75.6 & 58.4 & 76.5 & 69.5 & 60.1 & 70.2 & 56.9 & \underline{39.6} & \underline{8.98} \\
\midrule
\multirow{3}{*}{\makecell[l]{Mixture \\w/ MM-Bal.}}
& ep1 & 74.7 & 59.6 & 76.0 & 71.2 & 59.0 & 73.1 & 58.7 & 35.5 & 9.27 \\
& ep2 & \underline{\textbf{77.8}} & \underline{\textbf{61.7}} & \underline{\textbf{77.2}} & \underline{\textbf{71.8}} & 60.5 & \underline{74.8} & 58.5 & 41.2 & 8.31 \\
& ep3 & 77.1 & 60.5 & 77.0 & 69.8 & \underline{60.7} & 74.6 & \underline{\textbf{60.2}} & \underline{\textbf{44.1}} & \underline{\textbf{8.04}} \\

\bottomrule
\end{tabular}
\label{tab-multi_task_balanced_ablation}
\end{table}

\subsubsection{Dynamic Adaptive Balance Strategy}

We conducted a evaluation of our dynamic adaptive balance strategy across text-image, video, and audio modalities using constrained datasets. The evaluation was conducted on benchmark datasets specific to each modality: for text-image tasks, MMbench~\cite{liu2025mmbench}, OK-VQA~\cite{marino2019ok}, VQAv2~\cite{goyal2017making}, TextVQA~\cite{singh2019towards}, and GQA~\cite{hudson2019gqa} were employed; for video, MSVD-QA~\cite{xu2017video} and MSRVTT-QA~\cite{xu2017video} benchmarks were utilized; and for audio, we assessed performance on the AudioCaps~\cite{kim2019audiocaps} (AAC) and MLS~\cite{Pratap2020MLSAL}-English (ASR) tasks. The experimental outcomes are detailed in Table~\ref{tab-multi_task_balanced_ablation}.

In contrast to actual training pipeline, our evaluation involved instruction tuning starting from pre-trained models. Specifically, for each modality, we initially trained single-modality baseline models (the 'Sinle-modal Baselines' in Table~\ref{tab-multi_task_balanced_ablation}) individually over three epochs to establish the maximum achievable performance per modality. The results indicate that optimal performance was predominantly observed by the second epoch. However, the ASR task, due to its more complex patterns, had not fully converged even by the third epoch. Subsequently, we combined data from all three modalities to train a unified model (the 'Mixture w/o MM-Bal.' in Table~\ref{tab-multi_task_balanced_ablation}). Under this multimodal training regimen, the image-text modality reached its optimal performance at the second epoch, while the video modality achieved peak performance as early as the first epoch and with performance consistently decreasing in subsequent epochs. In contrast, the audio modality demonstrated continuous improvement, attaining its best performance by the third epoch. These observations underscore the imbalance in training progress among different modalities when engaged in multimodal training.

To address this imbalance, we introduced the dynamic adaptive balance strategy within our M2-omni training framework. This strategy dynamically adjusts the loss weights for each modality based on their respective training progress. In the context of this evaluation, it accelerates the training of the audio modality while appropriately reducing the learning weights for the image-text and video modalities to prevent overfitting. The evaluation results for this balanced training approach are denoted as 'Mixture w/ MM-Bal.' in Table~\ref{tab-multi_task_balanced_ablation}. The results demonstrate that, although some degree of imbalance among modalities persists, the balanced training strategy significantly alleviates the issues observed with simple mixed training: optimal performances across benchmarks are now concentrated around the second and third epochs, and performance across all modalities has been markedly enhanced. Moreover, under the balanced training strategy, the model achieved single-modality optimal performance in 7 out of 9 benchmarks. The best-performing model (at epoch 2) surpassed the optimal performance of each single-modality baseline in 6 out of 9 benchmarks (MMBench, OK-VQA, VQAv2, TextVQA, AudioCaps, MLS-English). Additionally, for the audio modality, the model at epoch 3 outperformed the single-modality baselines in 5 out of 9 benchmarks (OK-VQA, VQAv2, MSRVTT-QA, AudioCaps, MLS-English), with significant improvements in audio performance. These experimental results highlight the effectiveness of our dynamic adaptive balance strategy.

%% file: sections/5.conclusion.tex
\section{Conclusion}\label{sec:conclusion}

In this study, we propose M2-omni, a highly competitive omni-MLLM model to GPT-4o, characterized by its comprehensive modality and task support, as well as its exceptional performance. M2-omni demonstrates competitive performance across a diverse range of tasks, including image understanding, video understanding, interleaved image-text understanding, audio understanding and generation, as well as free-form image generation. We employ a multi-stage training approach to train M2-omni, which enables progressive modality alignment. To address the challenge of maintaining consistent performance across all modalities, we propose a step-wise balance strategy for pretraining and a dynamically adaptive balance strategy for instruction tuning, which can effectively mitigate the impact of significant variations in data volume and convergence rates across heterogeneous multimodal tasks. We publicly release M2-omni, along with its comprehensive training details, including data configurations and training procedures, to facilitate future research in this domain.

%% file: sections/6.ref.tex

{
\small
\bibliographystyle{plainnat}
\bibliography{ref}
}

%% file: sections/7.appendix.tex
\renewcommand\thefigure{S\arabic{figure}}
\renewcommand\thetable{S\arabic{table}}
\renewcommand\theequation{S\arabic{equation}}
\renewcommand\thealgorithm{S\arabic{algorithm}}
\setcounter{equation}{0}
\setcounter{table}{0}
\setcounter{figure}{0}
\setcounter{section}{0}
\setcounter{algorithm}{0}
\renewcommand\thesection{\Alph{section}}
\captionsetup[subfigure]{skip=0pt}

\section*{Appendix}

\section{More Qualitative Results}\label{subsec:appendix_cases}

In this section, we show examples of different modality tasks to demonstrate the capabilities of our \method model.

\begin{figure}[h]
    \centering
    \includegraphics[width=1.0\linewidth]{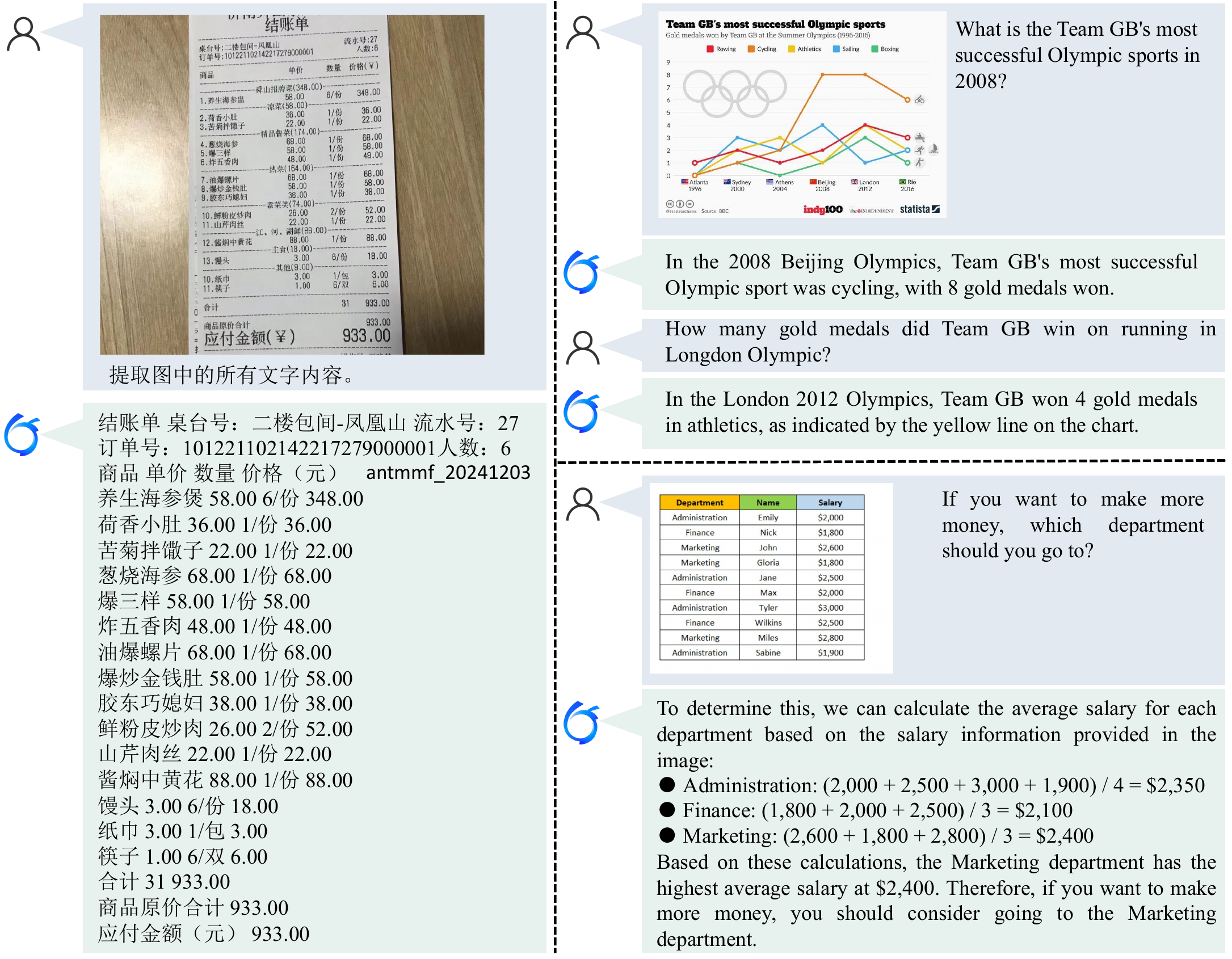}
    \caption{
    \textbf{Cases for complex OCR.}
    }
    \label{fig-exp_case_it_ocrs}
\end{figure}

\begin{figure}[h]
    \centering
    \begin{subfigure}{0.95\textwidth}
        \centering
        \includegraphics[width=\textwidth]{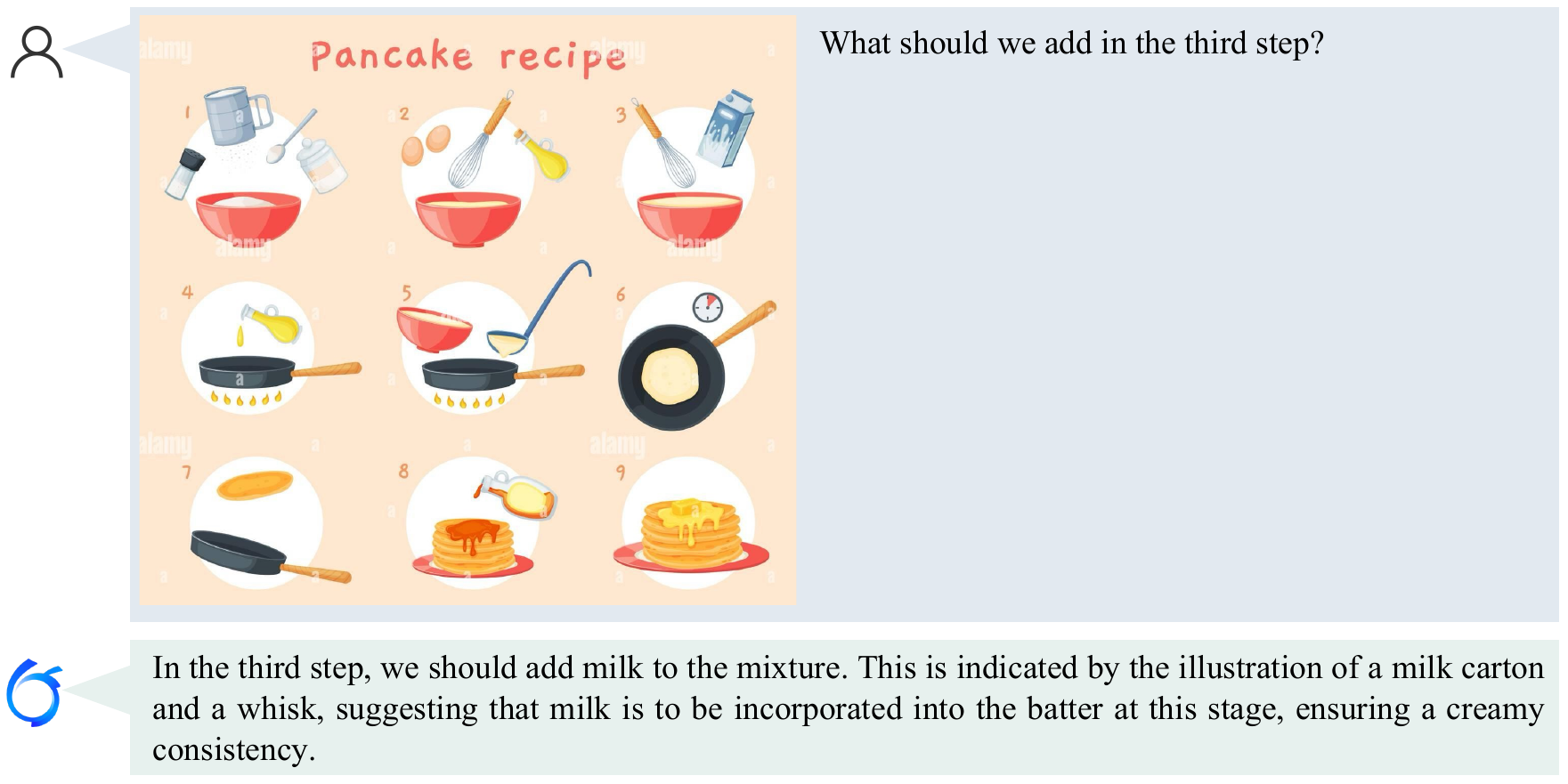}
        \vspace{-8pt}
        \caption{Flow chart understanding}
        \label{fig:case_recipe}
    \end{subfigure}
    \begin{subfigure}{0.95\textwidth}
        \centering
        \includegraphics[width=\textwidth]{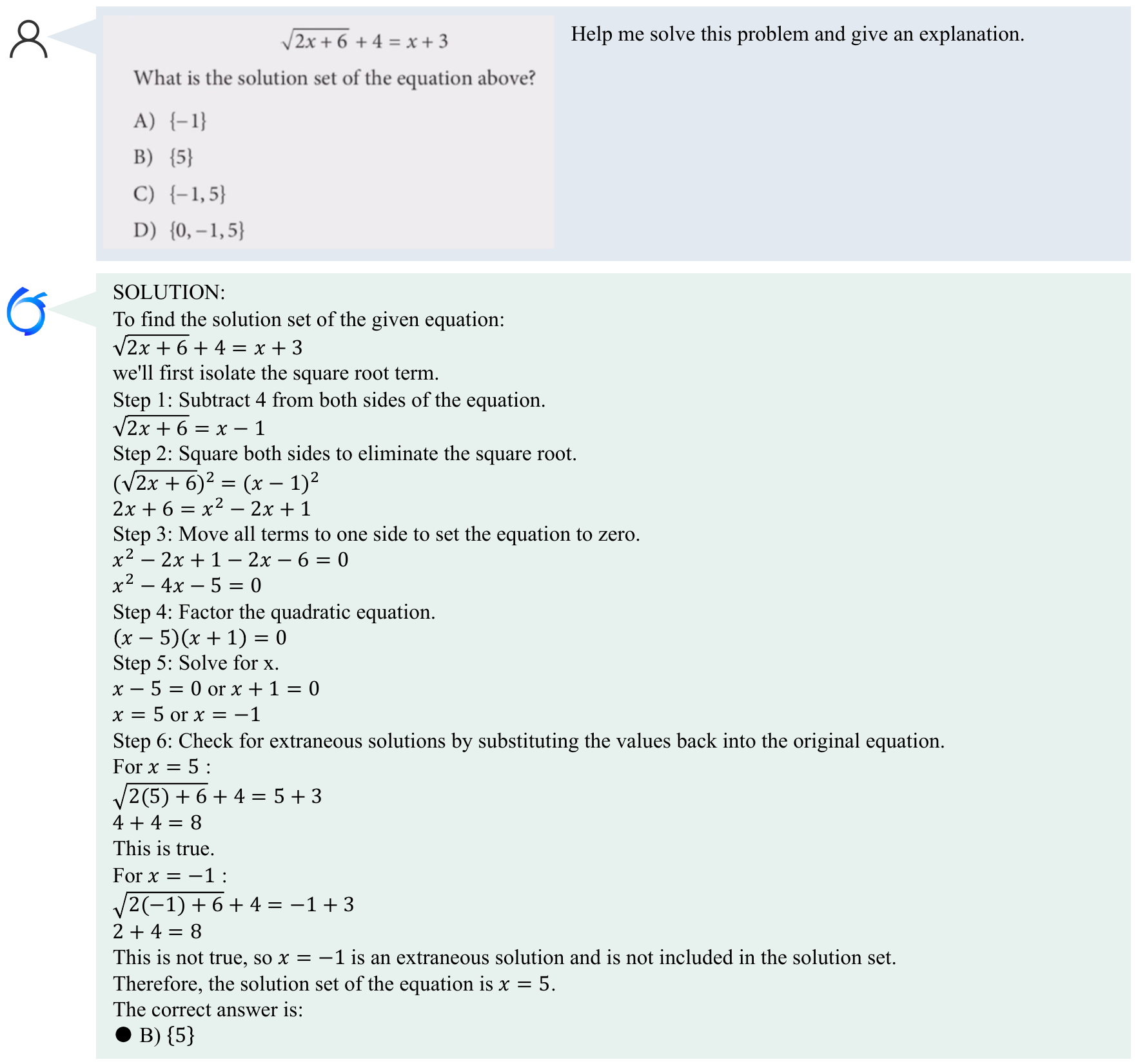}
        \vspace{-8pt}
        \caption{Mathematical problem solving}
        \label{fig:case_math}
    \end{subfigure}
    \caption{
    \textbf{Cases for information extraction and visual reasoning.}
    }
    \label{fig-exp_cases_it_reasoning}
\end{figure}

\textbf{Image Understanding.}
As shown in \cref{fig-exp_case_it_ocrs}, our \method model demonstrates an adeptness in complex OCR and reasoning.
And \cref{fig-exp_cases_it_reasoning}, the model displays its ability in flow chart understanding, and mathematical problem solving.
These cases demonstrate the model’s ability to handle various complex image-text understanding tasks.

\begin{figure}[h]
    \centering
    \begin{subfigure}{0.95\textwidth}
        \centering
        \includegraphics[width=\textwidth]{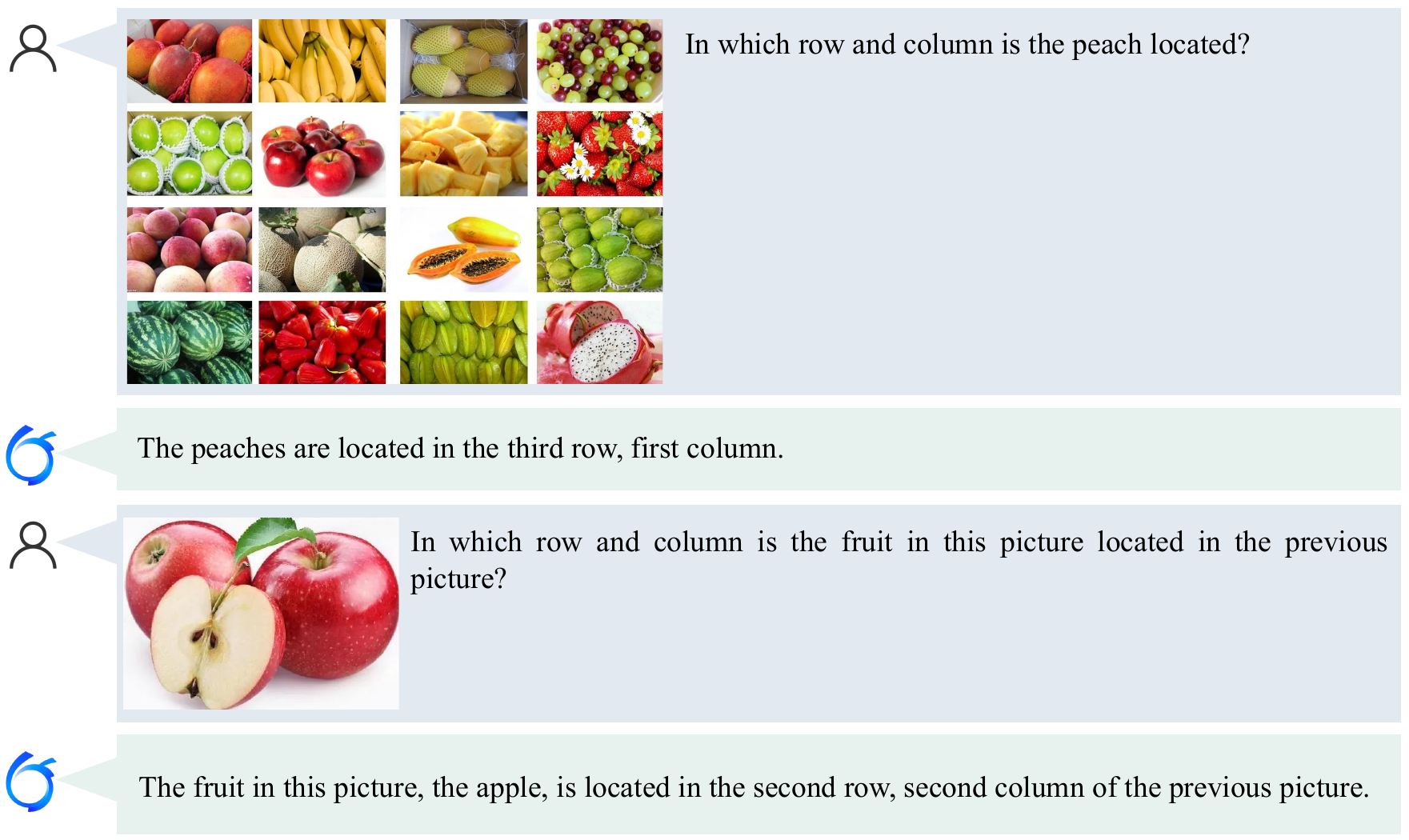}
        \vspace{-8pt}
        \caption{Multi-turn dialogue}
        \label{fig:case_interleave3}
    \end{subfigure}
    \begin{subfigure}{0.95\textwidth}
        \centering
        \includegraphics[width=\textwidth]{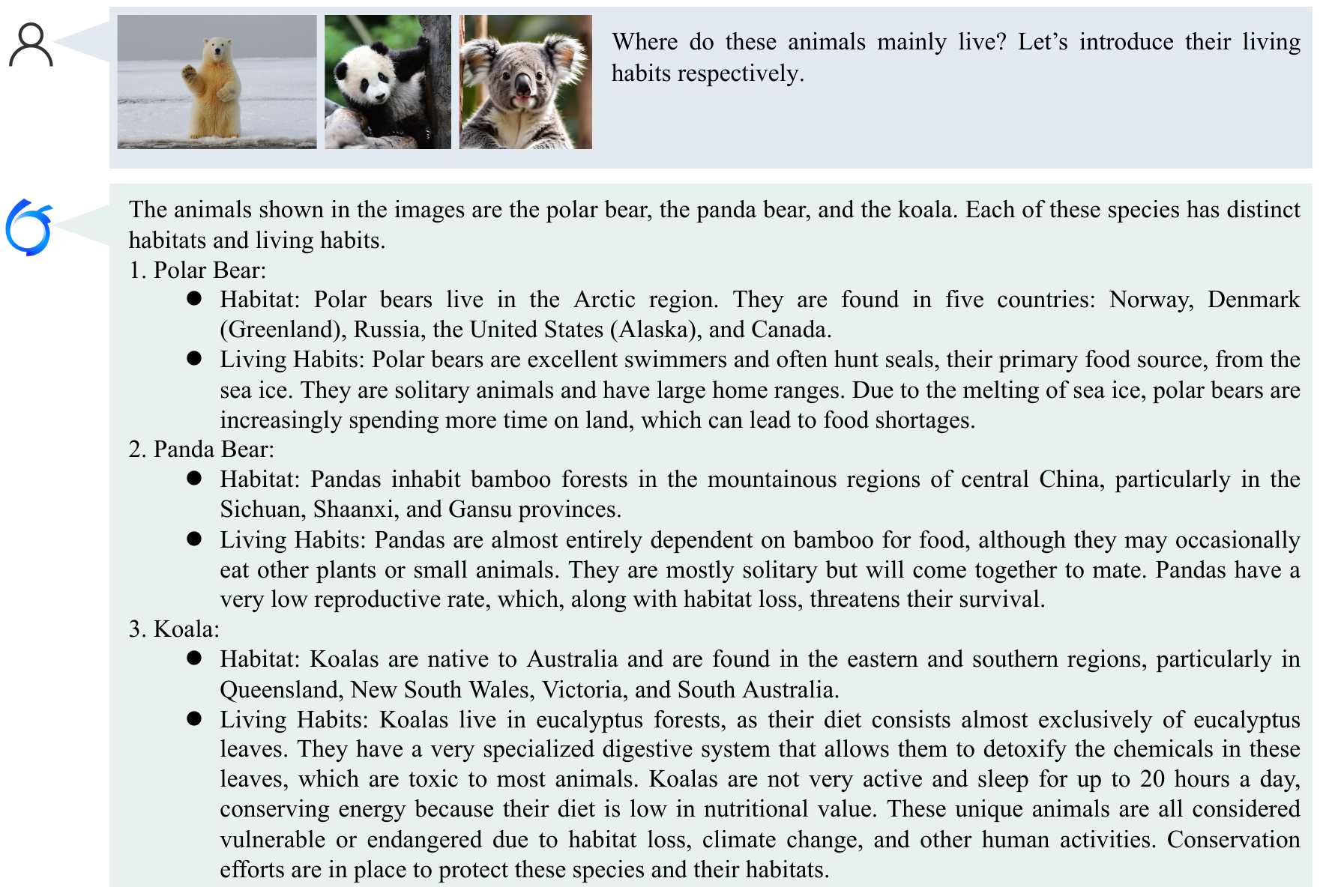}
        \vspace{-8pt}
        \caption{Multi-image understanding}
        \label{fig:case_interleave2}
    \end{subfigure}
    \caption{
    \textbf{Cases for interleaved image-text understanding.}
    }
    \label{fig-exp_case_it_interleave}
\end{figure}

\textbf{Interleave Image-Text Understanding.}
As shown in \cref{fig-exp_case_it_interleave}, the model shows its excellent ability in processing interleaved images-text.
The model has the ability to synthesize and understand the content of multiple images.

\begin{figure}[h]
    \centering
    \includegraphics[width=0.95\linewidth]{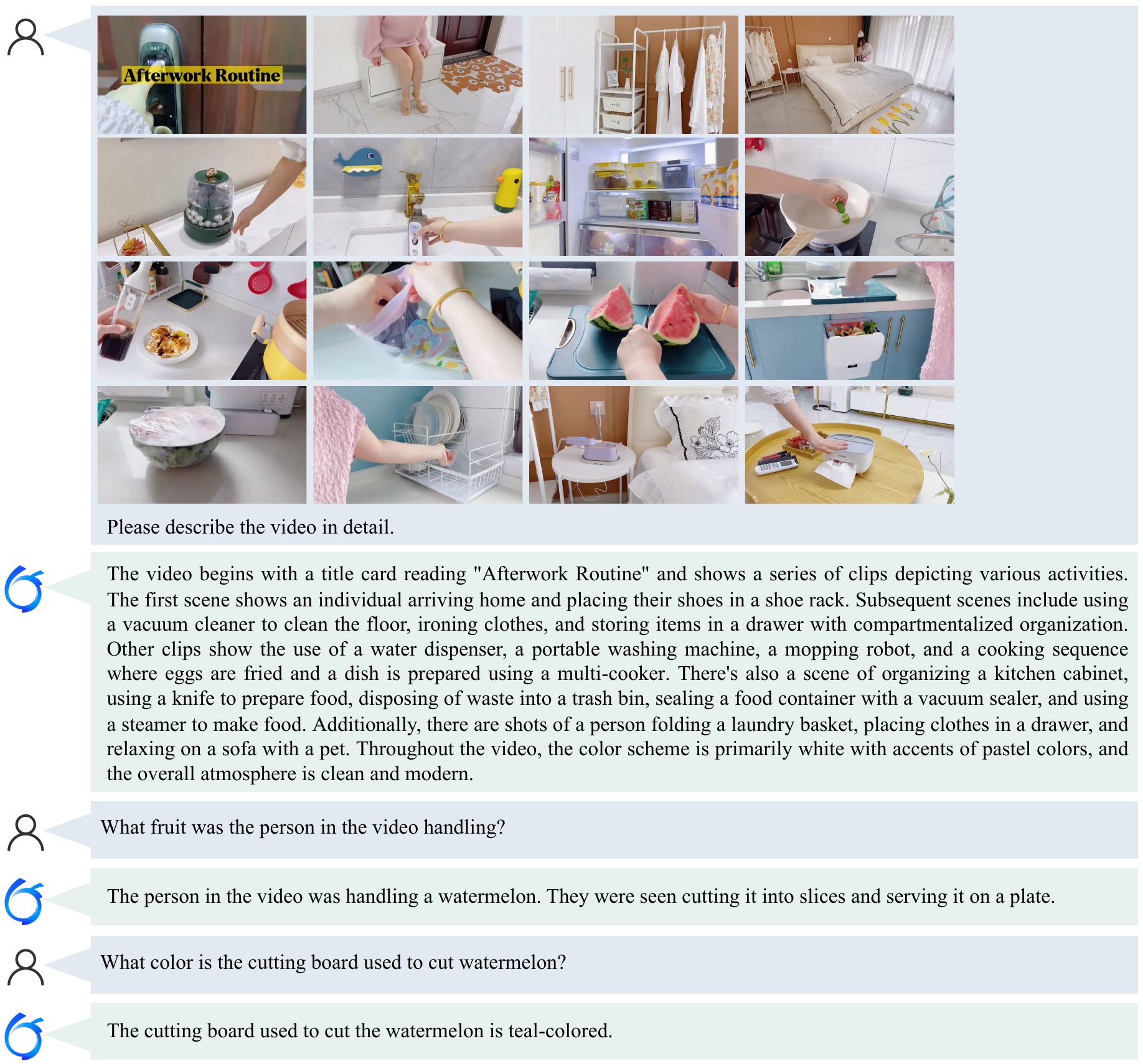}
    \caption{
    \textbf{Case for detailed video understanding.}
    }
    \label{fig-exp_case_video1}
\end{figure}

\textbf{Video Understanding.}
\cref{fig-exp_case_video1} displays several cases for video understanding.
We can see that our model can accurately capture the details in the video and understand the events in videos.

\begin{figure}[h]
    \centering
    \begin{subfigure}{0.95\textwidth}
        \centering
        \includegraphics[width=\textwidth]{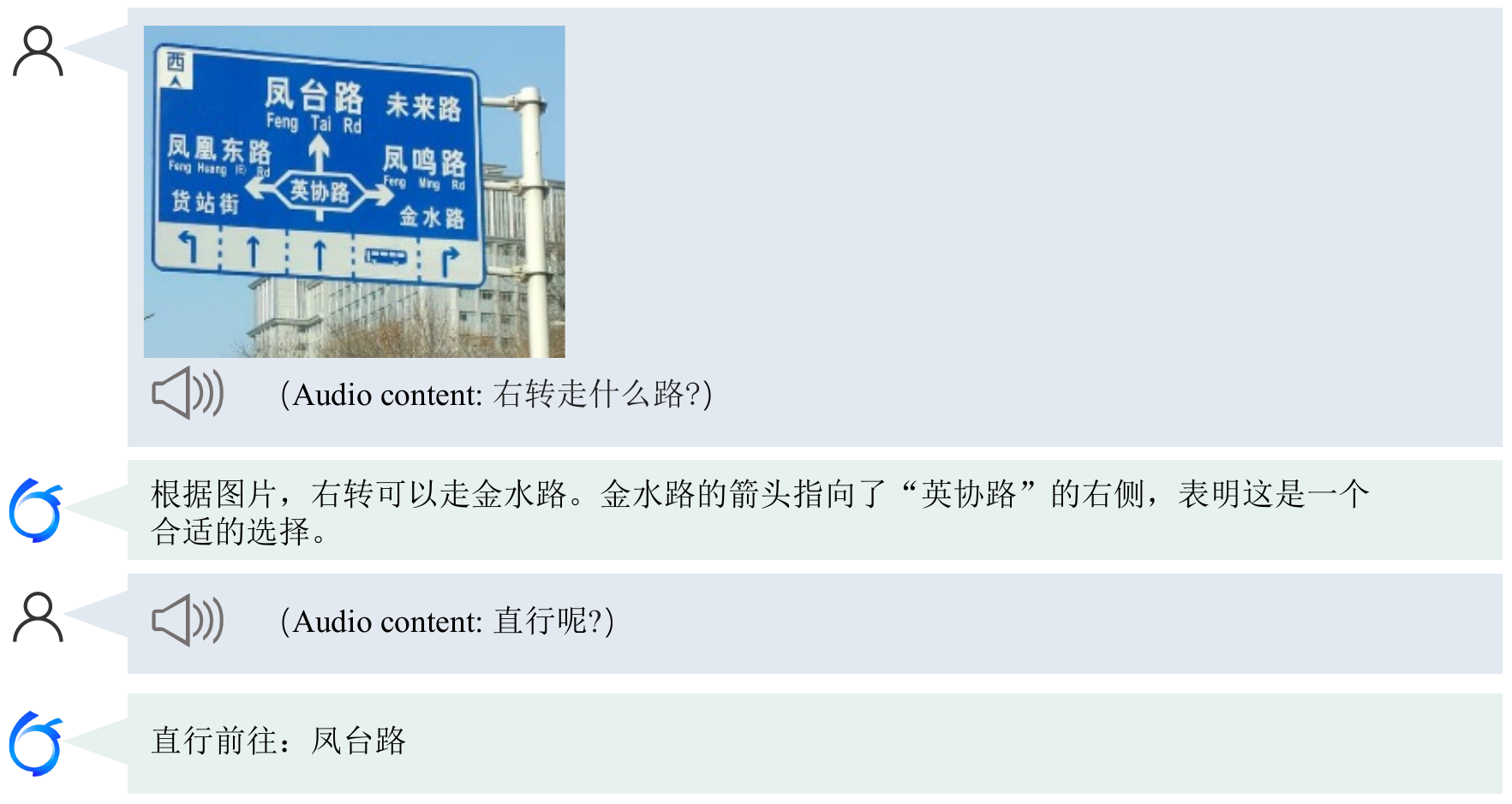}
        \vspace{-8pt}
        \caption{Image audio QA}
        \label{fig:case_image_audioqa}
    \end{subfigure}
    \begin{subfigure}{0.95\textwidth}
        \centering
        \includegraphics[width=\textwidth]{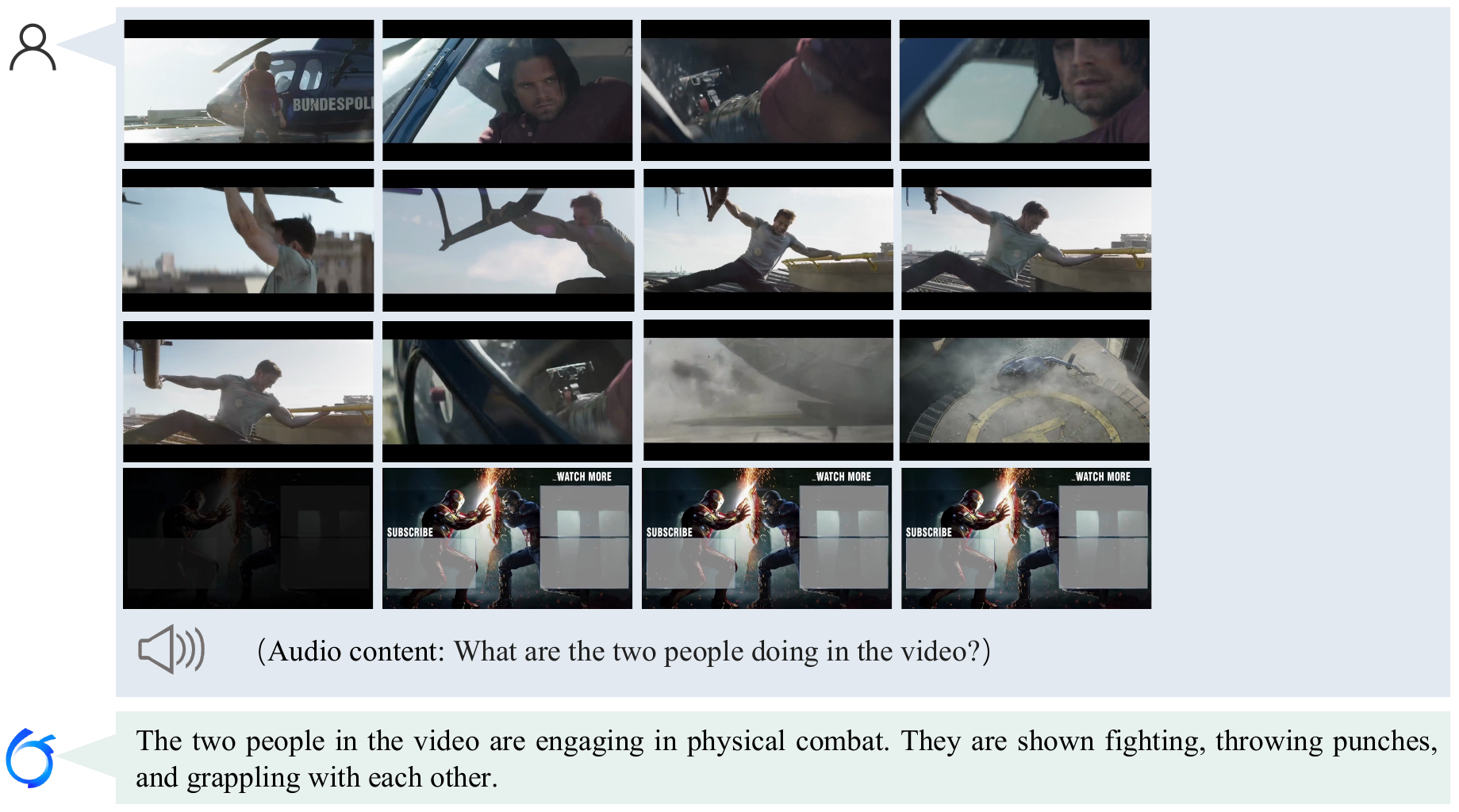}
        \vspace{-8pt}
        \caption{Video audio QA}
        \label{fig:case_video_audioqa}
    \end{subfigure}
    \caption{
    \textbf{Cases for audio QA.}
    }
    \label{fig-exp_case_audio_qa}
\end{figure}

\textbf{Audio QA.}
As shown in \cref{fig-exp_case_audio_qa}, the model can accurately understand the content of the audio and answer questions combined with the image/video content, which demonstrates the model's multimodal understanding ability.

\section{Training Data Details}\label{subsec:appendix_data}

In this section, we list the training data in detail, including data sources, sampling ratios, data volumes, and language information.

\begin{table}[h]
\centering
\caption{\textbf{Details of open source data in pre-training stage.}
}
\label{tab:appendix_pretrain_data}
\setlength{\tabcolsep}{7pt}
\begin{tabular}{l|l|l|l|l}
\hline
Dataset & Size & Ratio (\%) & Class & Language \\
\hline
Laion 5B~\cite{schuhmann2022laion} & 1555342102 & 67.10 & Image-Text & English \\
Zero~\cite{xiezero} & 151766788 & 6.55& Image-Text & Chinese \\
COYO~\cite{byeon2022coyo} & 61915700 &2.67& Image-Text & English \\
SBU~\cite{ordonez2011im2text} & 777671 & 0.03& Image-Text & English \\
CC3M~\cite{sharma2018conceptual} & 3105571 &0.13 & Image-Text & English \\
CC12M~\cite{changpinyo2021conceptual} & 10337890 & 0.45& Image-Text & English \\
Object365~\cite{shao2019objects365} & 1062993 & 0.05& Image-Text & English \\
Datacomp~\cite{gadre2024datacomp} & 187843384 & 8.10& Image-Text & Chinese \\
Clotho~\cite{Clotho} & 14465 & 0.00& Audio-Text & English \\
AudioCaps~\cite{AudioCaps} & 50939 &0.00 & Audio-Text & English \\
Libriheavy~\cite{Libriheavy} & 10454844 &0.45& Audio-Text & English \\
MACS~\cite{MACS} & 17275 & 0.00& Audio-Text & English \\
AudioSet~\cite{AudioSet} & 1909428 & 0.08& Audio-Text & English \\
VGGSound~\cite{VGGSound} & 95495 & 0.00& Audio-Text & English \\
AISHELL1~\cite{AISHELL1} & 134424 &0.01& Audio-Text & Chinese \\
WenetSpeech~\cite{WenetSpeech} & 14625245 & 0.63& Audio-Text & Chinese \\
GigaSpeech~\cite{GigaSpeech} & 8282988 & 0.36& Audio-Text & English \\
Librispeech~\cite{Librispeech} & 286652 &0.01 & Audio-Text & English \\
CoVoST2\_en\_en~\cite{CoVoST2} & 304937 & 0.01& Audio-Text & English \\
CoVoST2\_en\_zh-CN~\cite{CoVoST2} & 304937 &0.01 & Audio-Text & Chinese \\
CoVoST2\_zh-CN\_en~\cite{CoVoST2} & 11928 & 0.00& Audio-Text & English \\
CoVoST2\_zh-CN\_zh-CN~\cite{CoVoST2} & 11928 & 0.00& Audio-Text & Chinese \\
WavCaps~\cite{WavCaps} & 275282 & 0.01& Audio-Text & English \\
SPGISpeech~\cite{SPGISpeech} & 1966109 &0.08 & Audio-Text & English \\
KeSpeech~\cite{KeSpeech} & 973583 & 0.04& Audio-Text & Chinese \\
AliMeeting~\cite{AliMeeting} & 372886 &0.02& Audio-Text & Chinese \\
magicdata\_755h~\cite{MagicData_RAMC} & 584976 &0.03& Audio-Text & Chinese \\
MagicData\_RAMC~\cite{MagicData_RAMC} & 164972 &0.01 & Audio-Text & Chinese \\
Primewords\_100h~\cite{Primewords_100h} & 50383 & 0.00& Audio-Text & Chinese \\
FreeST~\cite{FreeST} & 102600 & 0.00& Audio-Text & Chinese \\
tal\_zh100h~\cite{TAL} & 93963 & 0.00& Audio-Text & Chinese \\
tal\_zhen587h~\cite{TAL} & 354680 & 0.02& Audio-Text & Chinese \\
aidatatang\_200zh~\cite{aidatatang_200zh} & 189121 & 0.01& Audio-Text & Chinese \\
TED\_LIUM\_v2~\cite{TED_LIUM} & 93437 & 0.00& Audio-Text & English \\
SlideSpeech~\cite{SlideSpeech} & 483731 & 0.02& Audio-Text & English \\
MMC4~\cite{zhu2024multimodal} & 36870000 & 1.59& Interleaved Image-Text & English \\
WuDao~\cite{yuan2021wudaocorpora} & 49896202 & 2.15& Text & Chinese \\
Pile~\cite{gao2020pile} & 136514954 & 5.89& Text & English \\
WebVid~\cite{bain2021frozen} & 1032719 & 0.04& Video-Text & English \\
Youku-mPLUG~\cite{xu2023youku} & 1098244 & 0.05& Video-Text & Chinese \\
CDIP-OCR~\cite{soboroff2022complex} & 9715288 & 0.42& OCR & English \\
OCR-IDL~\cite{biten2022ocr} & 26473994 & 1.14& OCR & English \\
Webvicob~\cite{kim2023web} & 27179852 & 1.17& OCR & English \\
WordScape~\cite{weber2023wordscape} & 8492083 & 0.37& OCR & English \\
MARIO-OCR~\cite{chen2024textdiffuser} & 6172608 & 0.27& OCR & English \\
\hline
\end{tabular}
\end{table}

\begin{table}[h]
\centering
\caption{\textbf{Details of open source image-text data in IT-Stage-1-Data (Part-1).}
}
\label{tab:appendix_sft_s1_it_1}
\setlength{\tabcolsep}{7pt}
\begin{tabular}{l|l|l|l|l}
\toprule
Dataset                         & Size    & Ratio (\%) & Data Class    & Language        \\
\midrule
TextCaps~\cite{sidorov2020textcaps}                        & 21942   & 0.08     & Captioning   & English         \\
LRV-Caption~\cite{liu2023aligning}                     & 254897  & 0.98     & Captioning   & English/Chinese \\
ShareGPT4V~\cite{chen2023sharegpt4v}            & 101178  & 0.39     & Captioning   & English         \\
Stanford40 Action~\cite{yao2011human}               & 9397    & 0.04     & Captioning   & English         \\
PixMo-CapQA~\cite{deitke2024molmo}                     & 229781  & 0.88     & Captioning   & English         \\
ChartQA~\cite{masry-etal-2022-chartqa}                         & 7548    & 0.03     & Chart        & English         \\
ChartBench~\cite{ChartBench}                      & 599616  & 2.29     & Chart        & English         \\
MMC-Inst~\cite{liu2023mmc}                        & 371027  & 1.42     & Chart        & English         \\
PlotQA~\cite{Methani_2020_WACV}                          & 157045  & 0.60      & Chart        & English         \\
FigureQA~\cite{kahou2017figureqa}                        & 100000  & 0.38     & Chart        & English         \\
DVQA~\cite{kafle2018dvqa}                            & 199364  & 0.76     & Chart        & English         \\
CRPE~\cite{wang2023allseeing}                            & 12896    & 0.05     & Chart        & English         \\
ChartGemma~\cite{masry2024chartgemma}                      & 163240  & 0.62     & Chart        & English         \\
TabMWP~\cite{lu2023dynamic}                          & 22662   & 0.09     & Chart        & English         \\
Chart2Text~\cite{kantharaj2022chart}                      & 27794   & 0.11     & Chart        & English         \\
LRV-Instruction-Chart~\cite{liu2023aligning}           & 21103   & 0.08     & Chart        & English         \\
ALLaVA~\cite{chen2024allava}                          & 977493  & 3.73     & Conversation & English/Chinese \\
ComVint~\cite{du2023makes}                         & 23586   & 0.11     & Conversation & English/Chinese \\
TallyQA~\cite{acharya2019tallyqa}                         & 249274  & 0.95     & Counting     & English         \\
Locount~\cite{Cai2020Locount}                         & 12660   & 0.05     & Counting     & English         \\
DocVQA~\cite{mathew2020docvqa}                          & 46191   & 0.17     & Document     & English         \\
DeepForm-VQA~\cite{aggarwal2023dublin}                    & 5367    & 0.02     & Document     & English         \\
Open-WikiTable~\cite{kweon2023open}                      & 2033    & 0.01     & Document     & English         \\
TabFact~\cite{2019TabFactA}                         & 12890   & 0.05     & Document     & English         \\
OCR-IDL~\cite{biten2022ocr}                         & 84183   & 0.32     & Document     & English         \\
kleister-charity~\cite{stanislawek2021kleister}                & 5175    & 0.02     & Document     & English         \\
DocReason25K~\cite{hu2024docowl}                    & 22568   & 0.09     & Document     & English         \\
HierText~\cite{long2022towards}                        & 8250    & 0.03     & Document     & English         \\
VisualMRC~\cite{VisualMRC2021}                       & 4794    & 0.02     & Document     & English         \\
DocStruct4M~\cite{hu2024docowl}                     & 299214  & 1.14     & Document     & English         \\
PubTabNet~\cite{zhong2019image}                       & 387665  & 1.48     & Document     & English         \\
arxiv\_qa~\cite{li2024multimodalarxiv}                       & 99992   & 0.38     & Document     & English         \\
PixMo-Docs~\cite{deitke2024molmo}                      & 255261  & 0.98     & Document     & English         \\
TAT-QA~\cite{zhu-etal-2021-tat}                          & 6630    & 0.02     & Document     & English         \\
Table-VQA~\cite{kim2024tablevqabench}                       & 84317   & 0.32     & Document     & English         \\
PixMo-AskModelAnything~\cite{deitke2024molmo}          & 157540  & 0.60      & General QA   & English         \\
COCO-VQA~\cite{ren2015exploring}                        & 28054   & 0.11     & General QA   & English         \\
VG-VQA~\cite{reich2024uncoveringpotentialvisualgrounding}                          & 14136   & 0.03     & General QA   & English/Chinese \\
SVIT~\cite{zhao2023svit}                            & 19818   & 0.08     & General QA   & English         \\
LVIS-Instruct4v~\cite{wang2023instruct4v}                 & 111169  & 0.42     & General QA   & Chinese         \\
RLAIF-V~\cite{yu2024rlaifv}                          & 22911   & 0.09     & General QA   & English         \\
RAVEN~\cite{zhang2019raven}                           & 41995   & 0.16     &  General QA    & English         \\
OK-VQA~\cite{marino2019ok}                          & 26994    & 0.10     & General QA   & English         \\
VQAv2~\cite{goyal2017making}                           & 249274  & 0.95     & General QA   & English         \\
GQA~\cite{hudson2019gqa}                             & 72148   & 0.28     & General QA   & English         \\
Q-Instruct~\cite{wu2023qinstruct}                      & 200534  & 0.77     & General QA   & English         \\
VSR~\cite{Liu2022VisualSR}                             & 9516    & 0.04     & General QA   & English         \\
VisCoT~\cite{shao2024visual}                          & 97490   & 0.37     & General QA   & English         \\
CogVLM-SFT-311K~\cite{wang2023cogvlm}                 & 296629  & 0.29     & General QA   & English/Chinese \\
ShareGPT4o~\cite{internvl_2024}                      & 96825   & 0.37     & General QA   & English         \\
image-textualization~\cite{pi2024image}            & 99573   & 0.38     & General QA   & English         \\
MapQA~\cite{chang2022mapqa}                           & 242501  & 0.93     & General QA   & English         \\
VizWiz~\cite{2018VizWiz}                          & 20523   & 0.08     & General QA   & English         \\
PathVQA~\cite{he2020pathvqa}                         & 32632   & 0.12     & General QA   & English         \\
ALFWorld~\cite{ALFWorld20}                        & 44968   & 0.17     & General QA   & English         \\
\bottomrule
\end{tabular}
\end{table}

\begin{table}[h]
\centering
\caption{\textbf{Details of open source image-text data in IT-Stage-1-Data (Part-2).}
}
\label{tab:appendix_sft_s1_it_2}
\setlength{\tabcolsep}{7pt}
\begin{tabular}{l|l|l|l|l}
\toprule
Dataset                         & Size    & Ratio (\%) & Data Class    & Language        \\
\midrule
lnqa~\cite{2020Connecting}                            & 207821  & 0.79     & General QA   & English         \\
OODVQA~\cite{0Benchmarking}                          & 16532    & 0.06     & General QA   & English         \\
Cambrian-10M~\cite{tong2024cambrian1}                    & 1863834 & 7.12     & General QA   & English         \\
Align-Anything-TI2T-Instruction~\cite{ji2024align} & 39216   & 0.15     & General QA   & English         \\
LRV-Instruction~\cite{liu2023aligning}                 & 180722  & 0.69     & General QA   & English         \\
VLGuard~\cite{zong2023safety}                         & 10000    & 0.04     & General QA   & English         \\
HalluciDoctor~\cite{yu2023hallucidoctor}                   & 104518   & 0.39     & General QA   & English         \\
M-HalDetect~\cite{gunjal2024detecting}                     & 28268   & 0.10     & General QA   & English         \\
illusionVQA~\cite{shahgir2024illusionvqa}                     & 4362    & 0.02     & General QA   & English         \\
HaloQuest~\cite{wang2024haloquest}                       & 7686    & 0.03     & General QA   & English         \\
Infinity-MM~\cite{gu2024infinitymm}                     & 78284   & 2.92     & General QA   & English         \\
RefCOCO~\cite{kazemzadeh2014referitgame}                         & 200537  & 0.54     & Grounding    & English         \\
Objects365~\cite{2020Objects365}                      & 1038990 & 3.96     & Grounding    & English/Chinese \\
GPT4Gen-RD-BoxCoT~\cite{chen2023shikra}               & 11268    & 0.04     & Grounding    & English         \\
AOKVQA~\cite{schwenk2022okvqa}                          & 34112   & 0.13     & Knowledge    & English         \\
KVQA~\cite{shah2019kvqa}                            & 20806   & 0.08     & Knowledge    & English         \\
CLEVR~\cite{johnson2017clevr}                           & 70000   & 0.27     & Knowledge    & English         \\
Super-CLEVR~\cite{li2023super}                      & 56913   & 0.22     & Knowledge    & English         \\
ViQuAE~\cite{lerner2022viquae}                          & 3153    & 0.01        & Knowledge    & English         \\
Co-Instruct~\cite{wu2024openended}                     & 9699    & 0.04     & Knowledge    & English         \\
CLEVR-Math~\cite{lindstrom2022clevr}                      & 675284  & 2.58     & Mathematics  & English         \\
MathV360K~\cite{shi2024math}                       & 328339  & 1.25     & Mathematics  & English         \\
GeoGPT4V~\cite{cai2024geogpt4v}                        & 18174   & 0.07     & Mathematics  & English         \\
GeoQA~\cite{chen2022geoqa}                          & 72318   & 0.28     & Mathematics  & English         \\
TextVQA~\cite{singh2019towards}                         & 21953   & 0.08     & OCR          & English         \\
OCR-VQA~\cite{mishra2019ocr}                          & 207572  & 0.79     & OCR          & English         \\
LLaVAR~\cite{zhang2023llavar}                          & 19800   & 0.08     & OCR          & English         \\
ICDAR19\_LSVT~\cite{sun2019icdar}                   & 30000   & 0.11     & OCR          & Chinese         \\
ICDAR19\_ReCTS~\cite{zhang2019icdar}                  & 19825   & 0.08     & OCR          & Chinese         \\
ICDAR19\_ArT~\cite{chng2019icdar2019}                    & 9344    & 0.03     & OCR          & Chinese         \\
Docmatix~\cite{2024docmatrix}                        & 101806  & 0.39     & OCR          & English         \\
DT-VQA~\cite{zhang2024exploring}                          & 20780   & 0.08     & OCR          & English         \\
Clock~\cite{yang2022s}                           & 3678    & 0.01        & OCR          & English         \\
COCO-Text V2.0~\cite{veit2016cocotext}                  & 13049   & 0.05     & OCR          & Chinese         \\
infographicsVQA~\cite{mathew2021infographicvqa}                 & 12078    & 0.05     & OCR          & English         \\
WebSRC~\cite{chen-etal-2021-websrc}                          & 9098    & 0.03     & OCR          & English         \\
CTW~\cite{yuan2019ctw}                        & 25871   & 0.10      & OCR          & Chinese         \\
RCTW-17~\cite{shi2017icdar2017}                            & 9262    & 0.03     & OCR          & Chinese         \\
ST-VQA~\cite{biten2019scene}                          & 17731   & 0.07     & OCR          & English         \\
TextOCR~\cite{singh2021textocr}                         & 21750   & 0.08     & OCR          & Chinese         \\
SynthText~\cite{Gupta16}                  & 857773  & 3.28     & OCR          & Chinese         \\
DTNutrCap~\cite{zhang2024exploring}                       & 4498    & 0.02     & OCR          & English         \\
DTScene~\cite{zhang2024exploring}                         & 15534    & 0.06     & OCR          & English         \\
ScreenQA~\cite{baechler2024screenai}                        & 161522   & 0.60     & OCR          & English         \\
Uber-Text~\cite{UberText}                       & 13941   & 0.05     & OCR          & Chinese         \\
MathWriting~\cite{gervais2024mathwriting}                     & 62726   & 0.24     & OCR          & English         \\
HME~\cite{yuan2022syntax}                             & 74502   & 0.28     & OCR          & English         \\
ScienceQA~\cite{lu2022learn}                       & 10332   & 0.04     & Science      & English         \\
AI2D~\cite{kembhavi2016diagram}                            & 9930    & 0.04     & Science      & English         \\
TQA~\cite{kembhavi2017you}                             & 19161    & 0.07     & Science      & English         \\
IconQA~\cite{lu2021iconqa}                          & 21393    & 0.08     & Science      & English         \\
PMC-VQA~\cite{zhang2023pmcvqa}                         & 152603  & 0.58     & Science      & English         \\
VQA-RAD~\cite{lau2018dataset}                         & 5379    & 0.02     & Science   & English         \\
\bottomrule
\end{tabular}
\end{table}

\begin{table}[h]
\centering
\caption{\textbf{Details of open source text data in IT-Stage-1-Data.}
}
\label{tab:appendix_sft_s1_nlp}
\setlength{\tabcolsep}{7pt}
\begin{tabular}{l|l|l|l|l}
\toprule
Dataset                            & Size    & Ratio (\%) & Language & Data Class   \\
\midrule
Evol-Instruct-Code~\cite{luo2023wizardcoder}                 & 66862   & 0.52     & English  & Code         \\
UltraInteract\_sft~\cite{yuan2024advancing}                 & 288579  & 2.27     & English  & Code         \\
glaive-code-assistant-v1~\cite{glaive-code-assistant}           & 136109  & 1.07     & English  & Code         \\
glaive-code-assistant-v2~\cite{glaive-code-assistant}           & 215166  & 1.69     & English  & Code         \\
glaive-code-assistant-v3~\cite{glaive-code-assistant}           & 475192  & 3.78     & English  & Code         \\
Magicoder~\cite{wei2024magicoder}                          & 108063  & 0.85     & English  & Code         \\
ALLaVA~\cite{chen2024allava}                             & 143000  & 1.12     & English  & Conversation \\
alpaca-gpt4~\cite{peng2023instruction}                        & 103998   & 0.83     & English  & Conversation \\
alpaca-gpt4\_zh~\cite{peng2023instruction}                    & 47347   & 0.37     & Chinese  & Conversation \\
UltraFeedback~\cite{cui2023ultrafeedback}                      & 121832   & 0.97     & English  & Conversation \\
Cambrian-10M~\cite{tong2024cambrian1}                       & 1713307 & 13.45    & English  & General QA   \\
SlimOrca~\cite{SlimOrca}                           & 517982  & 4.07     & English  & General QA   \\
Align-Anything-Instruction-100K~\cite{ji2024align}    & 105333  & 0.83     & English  & General QA   \\
Align-Anything-Instruction-100K-zh~\cite{ji2024align} & 104550  & 0.82     & Chinese  & General QA   \\
WizardLM\_evol\_instruct-v2~\cite{xu2024wizardlm}        & 142999  & 1.12     & English  & knowledge    \\
databricks\_dolly\_15k~\cite{DatabricksBlog2023DollyV2}             & 15011   & 0.12     & English  & knowledge    \\
Magpie-Qwen2~\cite{xu2024magpie}                       & 1000000 & 7.85     & English  & knowledge    \\
Infinity-Instruct~\cite{InfinityInstruct2024}                  & 2116735 & 16.62    & English  & knowledge    \\
MathInstruct~\cite{yue2023mammoth}                       & 261629  & 2.05     & English  & Mathematics  \\
NuminaMath-Cot~\cite{numina_math_datasets}                     & 859296  & 6.74     & English  & Mathematics  \\
DART-Math-Hard~\cite{tong2024dartmath}                     & 551000  & 4.32     & English  & Mathematics  \\
orca-math-word-problems-200k~\cite{mitra2024orcamath}       & 200035  & 1.57     & English  & Mathematics  \\
MetaMathQA~\cite{yu2023metamath}                         & 395000  & 3.10      & English  & Mathematics  \\
OpenMathInstruct-v2~\cite{toshniwal2024openmath2}                & 1000000 & 7.85     & English  & Mathematics  \\
MathGLM~\cite{yang2023gpt}                            & 200488  & 1.57     & Chinese  & Mathematics  \\
arxiv-physics-instruct~\cite{phi-arxiv-physics-instruct}             & 276286  & 2.17     & English  & Science      \\
\bottomrule
\end{tabular}
\end{table}

\begin{table}[ht]
\centering
\caption{\textbf{Details of open source image-text data in IT-Stage-2-Data (Part-1).}
}
\label{tab:appendix_sft_s2_it_1}
\setlength{\tabcolsep}{7pt}
\begin{tabular}{l|l|l|l|l}
\toprule
Dataset                         & Size    & Ratio (\%) & Data Class    & Language        \\
\midrule
LRV-Caption~\cite{liu2023aligning}                 & 106239 & 0.62       & Captioning   & English         \\
ShareGPT4V~\cite{chen2023sharegpt4v}               & 64917  & 0.38       & Captioning   & Chinese         \\
CogVLM-SFT-311K-caption~\cite{wang2023cogvlm}      & 67389  & 0.39       & Captioning   & Chinese         \\
textocr\_gpt4o\_train~\cite{TextOCR-GPT4o}         & 100324 & 0.58       & Captioning   & English         \\
PixMo-CapQA~\cite{deitke2024molmo}                 & 229781 & 1.33       & Captioning   & English         \\
ChartQA~\cite{masry-etal-2022-chartqa}             & 20128  & 0.12       & Chart        & English         \\
CRPE~\cite{wang2023allseeing}                      & 51584  & 0.3        & Chart        & English         \\
ChartBench~\cite{ChartBench}                       & 599616 & 3.48       & Chart        & English         \\
FigureQA~\cite{kahou2017figureqa}                  & 100000 & 0.58       & Chart        & English         \\
PlotQA~\cite{Methani_2020_WACV}                  & 157045 & 0.91       & Chart        & English         \\
MMC-Inst~\cite{liu2023mmc}                         & 371027 & 2.15       & Chart        & English         \\
DVQA~\cite{kafle2018dvqa}                          & 199364 & 1.16       & Chart        & English         \\
ChartGemma~\cite{masry2024chartgemma}              & 163240 & 0.95       & Chart        & English         \\
TabMWP~\cite{lu2023dynamic}                        & 226620 & 1.31       & Chart        & English         \\
Chart2Text~\cite{kantharaj2022chart}               & 277940 & 1.61       & Chart        & English         \\
LRV-Instruction-Chart~\cite{liu2023aligning}       & 21103  & 0.12       & Chart        & English         \\
ALLaVA~\cite{chen2024allava}                       & 165387 & 0.96       & Conversation & English/Chinese \\
ComVint~\cite{du2023makes}                         & 23586  & 0.14       & Conversation & English/Chinese \\
TallyQA~\cite{acharya2019tallyqa}                  & 49854  & 0.29       & Counting     & English         \\
Locount~\cite{Cai2020Locount}                      & 12660  & 0.07       & Counting     & English         \\
DocVQA~\cite{mathew2020docvqa}                     & 123176 & 0.72       & Document     & English         \\
DeepForm-VQA~\cite{aggarwal2023dublin}             & 7156   & 0.04       & Document     & English         \\
TabFact~\cite{2019TabFactA}                        & 12890  & 0.07       & Document     & English         \\
HierText~\cite{long2022towards}                    & 8250   & 0.05       & Document     & English         \\
TAT-QA~\cite{zhu-etal-2021-tat}                    & 4420   & 0.03       & Document     & English         \\
VisualMRC~\cite{VisualMRC2021}                     & 4794   & 0.03       & Document     & English         \\
OCR-IDL~\cite{biten2022ocr}                        & 168366 & 0.98       & Document     & English         \\
DocReason25K~\cite{hu2024docowl}                   & 22568  & 0.13       & Document     & English         \\
DocStruct4M~\cite{hu2024docowl}                    & 89764  & 0.52       & Document     & English         \\
arxiv\_qa~\cite{li2024multimodalarxiv}        & 99992  & 0.58       & Document     & English         \\
PixMo-Docs~\cite{deitke2024molmo}                  & 255261 & 1.48       & Document     & English         \\
SVIT~\cite{zhao2023svit}                           & 39636  & 0.23       & General QA   & English         \\
GQA~\cite{hudson2019gqa}                           & 72148  & 0.42       & General QA   & English         \\
MapQA~\cite{chang2022mapqa}                        & 242501 & 1.41       & General QA   & English         \\
ShareGPT4o~\cite{internvl_2024}                   & 193650 & 1.12       & General QA   & English         \\
CogVLM-SFT-311K-caption~\cite{wang2023cogvlm}      & 935553 & 3.69       & General QA   & English/Chinese \\
image-textualization~\cite{pi2024image}            & 298719 & 1.73       & General QA   & English         \\
lnqa~\cite{2020Connecting}                         & 207821 & 1.21       & General QA   & English         \\
OODVQA~\cite{0Benchmarking}                        & 8266   & 0.05       & General QA   & English         \\
VizWiz~\cite{2018VizWiz}                           & 20523  & 0.12       & General QA   & English         \\
VQAv2~\cite{goyal2017making}                       & 249274 & 1.45       & General QA   & English         \\
RAVEN~\cite{zhang2019raven}                        & 83990  & 0.49       & General QA   & English         \\
RLAIF-V~\cite{yu2024rlaifv}                        & 45822  & 0.27       & General QA   & English         \\
VSR~\cite{Liu2022VisualSR}                         & 4758   & 0.03       & General QA   & English         \\
Cambrian-10M~\cite{tong2024cambrian1}              & 28054  & 0.16       & General QA   & English         \\
Align-Anything-TI2T-Instruction~\cite{ji2024align} & 78432  & 0.46       & General QA   & English         \\
LRV-Instruction~\cite{liu2023aligning}             & 361444 & 2.1        & General QA   & English/Chinese \\
VLGuard~\cite{zong2023safety}                      & 10000  & 0.06       & General QA   & English         \\
HalluciDoctor~\cite{yu2023hallucidoctor}           & 104528 & 0.61       & General QA   & English         \\
M-HalDetect~\cite{gunjal2024detecting}             & 28268  & 0.17       & General QA   & English         \\
illusionVQA~\cite{shahgir2024illusionvqa}          & 7270   & 0.04       & General QA   & English         \\
HaloQuest~\cite{wang2024haloquest}                 & 23058  & 0.13       & General QA   & English         \\
Infinity-MM~\cite{gu2024infinitymm}                & 782884 & 4.55       & General QA   & English         \\
PixMo-AskModelAnything~\cite{deitke2024molmo}      & 157540 & 0.91       & General QA   & English         \\
VisCoT~\cite{shao2024visual}                       & 175022 & 1.02       & General QA   & English         \\
DocStruct-Grounding~\cite{hu2024docowl}            & 99983  & 0.58       & Grounding    & English         \\
AOKVQA~\cite{schwenk2022okvqa}                     & 17056  & 0.1        & Knowledge    & English         \\
\bottomrule
\end{tabular}
\end{table}

\begin{table}[h]
\centering
\caption{\textbf{Details of open source image-text data in IT-Stage-2-Data (Part-2).}
}
\label{tab:appendix_sft_s2_it_2}
\setlength{\tabcolsep}{7pt}
\begin{tabular}{l|l|l|l|l}
\toprule
Dataset                         & Size    & Ratio (\%) & Data Class    & Language        \\
\midrule
ViQuAE~\cite{lerner2022viquae}                     & 2102   & 0.01       & Knowledge    & English         \\
KVQA~\cite{shah2019kvqa}                           & 41612  & 0.24       & Knowledge    & English         \\
Co-Instruct~\cite{wu2024openended}                 & 12932  & 0.08       & Knowledge    & English         \\
CLEVR~\cite{johnson2017clevr}                      & 70000  & 0.41       & Knowledge    & English         \\
Super-CLEVR~\cite{li2023super}                     & 56913  & 0.33       & Knowledge    & English         \\
GeoGPT4V~\cite{cai2024geogpt4v}                    & 36348  & 0.21       & Mathematics  & English         \\
MathV360K~\cite{shi2024math}                       & 328339 & 1.91       & Mathematics  & English         \\
CLEVR-Math~\cite{lindstrom2022clevr}               & 675284 & 3.92       & Mathematics  & English         \\
Geo170k~\cite{gao2023g}                            & 176602 & 1.02       & Mathematics  & English         \\
GeoQA~\cite{chen2022geoqa}                         & 144636 & 0.84       & Mathematics  & English         \\
TextVQA~\cite{singh2019towards}                    & 21953  & 0.13       & OCR          & English         \\
OCR-VQA~\cite{mishra2019ocr}                       & 207572 & 1.2        & OCR          & English         \\
ICDAR19\_LSVT~\cite{sun2019icdar}             & 30000  & 0.17       & OCR          & Chinese         \\
ICDAR19\_ReCTS~\cite{zhang2019icdar}          & 19825  & 0.12       & OCR          & Chinese         \\
ICDAR19\_ArT~\cite{chng2019icdar2019}         & 4672   & 0.03       & OCR          & Chinese         \\
Clock~\cite{yang2022s}                             & 4904   & 0.03       & OCR          & English         \\
DT-VQA~\cite{zhang2024exploring}                   & 83120  & 0.48       & OCR          & English         \\
Docmatix~\cite{2024docmatrix}              & 306376 & 1.78       & OCR          & English         \\
DTNutrCap~\cite{zhang2024exploring}                & 4498   & 0.03       & OCR          & English         \\
DTScene~\cite{zhang2024exploring}                  & 15534  & 0.09       & OCR          & English         \\
infographicsVQA~\cite{mathew2021infographicvqa}    & 16104  & 0.09       & OCR          & English         \\
WebSRC~\cite{chen-etal-2021-websrc}                & 18196  & 0.11       & OCR          & English         \\
TextOCR~\cite{singh2021textocr}                    & 87000  & 0.5        & OCR          & Chinese         \\
ST-VQA~\cite{biten2019scene}                       & 35462  & 0.21       & OCR          & English         \\
CTW~\cite{yuan2019ctw}                             & 51742  & 0.3        & OCR          & Chinese         \\
RCTW-17~\cite{shi2017icdar2017}                    & 9262   & 0.05       & OCR          & Chinese         \\
COCO-Text   V2.0~\cite{veit2016cocotext}           & 13049  & 0.08       & OCR          & Chinese         \\
SynthText~\cite{Gupta16}                           & 85777  & 0.5        & OCR          & Chinese         \\
ScreenQA~\cite{baechler2024screenai}               & 80761  & 0.47       & OCR          & English         \\
Uber-Text~\cite{UberText}                          & 27882  & 0.16       & OCR          & Chinese         \\
HME~\cite{yuan2022syntax}                          & 148984 & 0.86       & OCR          & English         \\
MathWriting~\cite{gervais2024mathwriting}          & 31363  & 0.18       & OCR          & English         \\
IconQA~\cite{lu2021iconqa}                         & 7131   & 0.04       & Science      & English         \\
TQA~\cite{kembhavi2017you}                         & 76644  & 0.44       & Science      & English         \\
AI2D~\cite{kembhavi2016diagram}                    & 39720  & 0.23       & Science      & English         \\
ScienceQA~\cite{lu2022learn}                       & 123984 & 0.72       & Science      & English         \\
VQA-RAD~\cite{lau2018dataset}                      & 10758  & 0.06       & Science      & English         \\
PMC-VQA~\cite{zhang2023pmcvqa}                     & 152603 & 0.89       & Science      & English          \\
\bottomrule
\end{tabular}
\end{table}

\begin{table}[h]
\centering
\caption{\textbf{Details of Interleaved image-text data in IT-Stage-2-Data.}}
\label{tab:multi_image_stage2}
\setlength{\tabcolsep}{7pt}
\begin{tabular}{l|l|l|l|l}
\hline
Dataset & Size & Ratio (\%) & Class & Language \\
\hline
InternVL-SA-1B-Caption-EN~\cite{chen2023internvl} & 77704 & 2.0 & Caption & English \\
InternVL-SA-1B-Caption-CN~\cite{chen2023internvl} & 77704 & 1.0 & Caption & Chinese \\
MMDU-EN~\cite{wang2024mmduet} & 74998 & 4.0 & General QA & English \\
MMDU-CN~\cite{wang2024mmduet} & 33882 & 2.0 & General QA & Chinese \\
M4-Instruct~\cite{li2024llavanext} & 615814 & 1.0 & General QA & English \\
\textcolor{gray}{Synthetic GEO-mv} & 241060 & 1.0 & General QA & English \\
\textcolor{gray}{Synthetic ScienceQA-mv} & 45916 & 2.0 & General QA & English \\
MP-DocVQA (en)~\cite{tito2023hier} & - &- & Document & English \\
MP-Docmatix~\cite{2024docmatrix} & - &- & Document & English \\
\hline
\end{tabular}
\end{table}

\begin{table}[h]
\centering
\caption{\textbf{Details of Video-text data in IT-Stage-2-Data.}}
\label{tab:video_text_stage2}
\setlength{\tabcolsep}{7pt}
\begin{tabular}{l|l|l|l|l}
\hline
Dataset & Size & Ratio (\%) & Class & Language \\
\hline
Vript~\cite{yang2024vript} & - & - & Caption & English \\
OpenVid~\cite{nan2024openvid} & - & - & Caption & English \\
Mementos~\cite{wang2024mementos} & - & - & Caption & English \\
ShareGPT4o-Video~\cite{chen2024far} & 2111 & 4.0 & Caption & English \& Chinese \\
ShareGPT4Video~\cite{chen2024sharegpt4video} & 255000 & 1.0 & Caption & English \\
VideoGPT+~\cite{Maaz2024VideoGPT+} & - & - & Caption & English \\
TextVR~\cite{wu2023largecrossmodal} & 39648 & 1.0 & Caption & English \\
\textcolor{gray}{Synthetic VideoDetailCaption} & 56748 & 4.0 & Caption & English \\
EgoTaskQA~\cite{jia2022egotaskqa} & 61710 & 1.0 & General QA & English \\
CLEVRER~\cite{yi2019clevrer} & 82620 & 1.0 & General QA & English \\
NExT-QA~\cite{xiao2021next} & 34132 & 1.0 & General QA & English \\
LLaVA-Video~\cite{zhang2024video} & 1367388 & 1.0 & General QA & English \\
FineVideo~\cite{FineVideo} & 87502 & 1.0 & General QA & English \\
VideoBlink (en)~\cite{videoblink} & 2590 & 4.0 & General QA & English \\
TGIF-QA~\cite{yuntgifqa} & 52696 & 1.0 & General QA & English \\
\hline
\end{tabular}
\end{table}

\begin{table}[h]
\centering
\caption{\textbf{Details of video-text data in IT-Stage-3-Data.}}
\label{tab:video_text_stage3}
\setlength{\tabcolsep}{7pt}
\begin{tabular}{l|l|l|l|l}
\hline
Dataset & Size & Ratio (\%) & Class & Language \\
\hline
TextVR~\cite{wu2023largecrossmodal} & 39648 & 1.0 & Caption & English \\
ShareGPT4o-Video~\cite{chen2024far} & 2111 & 4.0 & Caption & English \& Chinese \\
\textcolor{gray}{Synthetic VideoAction} & 7182 & 4.0 & Caption & English \\
\textcolor{gray}{Synthetic VideoDetailCaption} & 56748 & 4.0 & Caption & English \\
EgoTaskQA~\cite{jia2022egotaskqa} & 61710 & 2.0 & General QA & English \\
CLEVRER~\cite{yi2019clevrer} & 82620 & 1.0 & General QA & English \\
NExT-QA~\cite{xiao2021next} & 34132 & 1.0 & General QA & English \\
LLaVA-Video~\cite{zhang2024video} & 1367388 & 0.5 & General QA & English \\
FineVideo~\cite{FineVideo} & 87502 & 1.0 & General QA & English \\
VideoBlink (en)~\cite{videoblink} & 2590 & 2.0 & General QA & English \\
MoVid~\cite{chen2024motionllm} & 24699 & 2.0 & General QA & English \\
FunQA~\cite{xie2024funqa} & 21012 & 1.0 & General QA & English \\
TGIF-QA~\cite{yuntgifqa} & 52696 & 1.0 & General QA & English \\
AlignAnythingVideo~\cite{ji2024alignanything} & 9501 & 6 & General QA & English \\
STAR~\cite{wu2024starbench} & 45731 & 2.0 & General QA & English \\
MMWorld~\cite{he2024mmworld} & 1976 & 4.0 & General QA & English \\
IntentQA~\cite{Li0HF23} & 40320 & 2.0 & General QA & English \\
\hline
\end{tabular}
\end{table}

\begin{table}[h]
\centering
\caption{\textbf{Details of open source audio-text data in IT-Stage-3-Data.}
}
\label{tab:appendix_audio_text}
\setlength{\tabcolsep}{7pt}
\begin{tabular}{l|l|l|l|l}
\toprule
Dataset                         & Size    & Ratio (\%) & Data Class    & Language        \\
\midrule
Libriheavy~\cite{Libriheavy} & 10454844 & 24.77 &ASR &English \\
GigaSpeech~\cite{GigaSpeech} & 8282988 & 19.62 &ASR &English \\
Librispeech~\cite{Librispeech} & 286652 & 0.68 &ASR &English \\
CoVoST2\_en\_en~\cite{CoVoST2} & 304937 & 0.72 &ASR &English \\
SPGISpeech~\cite{SPGISpeech} & 1966109 & 4.66 &ASR &English \\
TEDLIUM\_v2~\cite{TED_LIUM} & 93437 & 0.22 &ASR &English \\
SlideSpeech~\cite{SlideSpeech} & 483731 & 1.15 &ASR &English \\
CoVoST2\_cn\_cn~\cite{CoVoST2} & 11928 & 0.03 &ASR &Chinese \\
AISHELL1~\cite{AISHELL1} & 134424 & 0.32 &ASR &Chinese \\
WenetSpeech~\cite{WenetSpeech} & 14625245 & 34.65 &ASR &Chinese \\
KeSpeech~\cite{KeSpeech} & 973583 & 2.31 &ASR &Chinese \\
AliMeeting~\cite{AliMeeting} & 372886 & 0.88 &ASR &Chinese \\
MagicData\_755h~\cite{MagicData_RAMC} & 584976 & 1.39 &ASR &Chinese \\
MagicData\_RAMC~\cite{MagicData_RAMC} & 164972 & 0.39 &ASR &Chinese \\
Primewords\_100h~\cite{Primewords_100h} & 50383 & 0.12 &ASR &Chinese \\
FreeST~\cite{FreeST} & 102600 & 0.24 &ASR &Chinese \\
Tal\_zh100h~\cite{TAL} & 93963 & 0.22 &ASR &Chinese \\
Tal\_zhen587h~\cite{TAL} & 354680 & 0.84 &ASR &Chinese \\
Aidatatang\_200zh~\cite{aidatatang_200zh} & 189121 & 0.45 &ASR &Chinese \\
CoVoST2\_en\_cn~\cite{CoVoST2} & 304937 & 0.72 &AST &Chinese \\
CoVoST2\_cn\_en~\cite{CoVoST2} & 11928 & 0.03 &AST &English \\
WavCaps~\cite{WavCaps} & 275282 & 0.65 &AAC &English \\
Clotho~\cite{Clotho} & 14465   & 0.03     &AAC    & English         \\
AudioCaps~\cite{AudioCaps} & 50939 & 0.12 &AAC &English \\
MACS~\cite{MACS} & 17275 & 0.04 &AAC &English \\
AudioSet~\cite{AudioSet} & 1909428 & 4.52 &AAT &English \\
VGGSound~\cite{VGGSound} & 95495 & 0.23 &AAT &English \\
\bottomrule
\end{tabular}
\end{table}

\begin{table}[h]
\centering
\caption{\textbf{Details of audio-QA data in IT-Stage-3-Data.}
}
\label{tab:appendix_sft_s3_audioqa}
\setlength{\tabcolsep}{8pt}
\begin{tabular}{l|l|l|l|l}
\toprule
Dataset   & Size    & Ratio (\%) & Data Class     & Language        \\
\midrule
COCO-VQA~\cite{ren2015exploring}  & 879598  & 11.81  & Image-Audio-QA & English/Chinese \\
VG-VQA~\cite{reich2024uncoveringpotentialvisualgrounding}    & 300000  & 4.03  & Image-Audio-QA & English/Chinese \\
SVIT~\cite{zhao2023svit}      & 40000   & 0.54  & Image-Audio-QA & English/Chinese \\
OK\_VQA~\cite{marino2019ok}   & 18018   & 0.24  & Image-Audio-QA & English/Chinese \\
VQAv2~\cite{goyal2017making}     & 887514  & 11.91  & Image-Audio-QA & English/Chinese \\
GQA~\cite{hudson2019gqa}       & 1886000 & 25.31  & Image-Audio-QA & English/Chinese \\
TextVQA~\cite{singh2019towards}   & 69204   & 0.93  & Image-Audio-QA & English/Chinese \\
TextCaps~\cite{sidorov2020textcaps}  & 219530  & 2.95   & Image-Audio-QA & English/Chinese \\
OCR-VQA~\cite{mishra2019ocr}   & 2004292 & 26.90  & Image-Audio-QA & English/Chinese \\
TextVR~\cite{wu2023largecrossmodal}    & 79296   & 1.06  & Video-Audio-QA & English/Chinese \\
VideoChat~\cite{li2023videochat} & 13778   & 0.18  & Video-Audio-QA & English/Chinese \\
WebVid~\cite{Bain21}    & 800000  & 10.74  & Video-Audio-QA & English/Chinese \\
NExT-QA~\cite{xiao2021next}   & 68264   & 0.92  & Video-Audio-QA & English/Chinese \\
CLEVRER~\cite{yi2019clevrer}   & 80000   & 1.07  & Video-Audio-QA & English/Chinese \\
TGIF-QA~\cite{yuntgifqa}   & 105392  & 1.41  & Video-Audio-QA & English/Chinese \\
\bottomrule
\end{tabular}
\end{table}